\documentclass[sigconf]{acmart}
\usepackage{amsfonts} 
\usepackage{nicefrac} 
\usepackage{microtype} 
\usepackage{amsthm}
\usepackage{amsmath}
\usepackage{algorithm}
\usepackage{algorithmic}
\usepackage{subfig}
\def \S {\mathbf{S}}

\def \x {\mathbf{x}}

\def \X {\mathcal{X}}

\def \R {\mathbb{R}}

\def \x {\mathbf{x}}

\def \E {\mathrm{E}}

\def \x {\mathbf{x}}
\def \p {\mathbf{p}}

\def \c {\mathbf{c}}

\def \P {\mathcal{P}}

\def \E {\mathrm{E}}
\def \x {\mathbf{x}}

\def \R {\mathbb{R}}
\def \S {\mathcal{S}}

\def \c {\mathbf{c}}

\def \p {\mathbf{p}}

\def \X {\mathcal{X}}

\def \P {\mathbb{P}}

\DeclareMathOperator*{\argmin}{arg\,min}

\AtBeginDocument{%
  \providecommand\BibTeX{{%
    \normalfont B\kern-0.5em{\scshape i\kern-0.25em b}\kern-0.8em\TeX}}}

\setcopyright{rightsretained}
\copyrightyear{2021}
\acmYear{2021}
\acmDOI{10.1145/3442188.3445878}

\acmConference[FAccT '21]{ACM Conference on Fairness, Accountability, and Transparency}{March 1--10, 2021}{Virtual Event, Canada}
\acmBooktitle{ACM Conference on Fairness, Accountability, and Transparency (FAccT '21), March 1--10, 2021, Virtual Event, Canada}\acmDOI{10.1145/3442188.3445878}
\acmISBN{978-1-4503-8309-7/21/03}

\begin{document}

\title{Towards Fair Deep Anomaly Detection}


\author{Hongjing Zhang}
\affiliation{%
  \institution{University of California, Davis}
  \postcode{95616}
}
\email{hjzzhang@ucdavis.edu}

\author{Ian Davidson}
\affiliation{%
  \institution{University of California, Davis}
  \postcode{95616}
}
\email{davidson@cs.ucdavis.edu}


\begin{abstract}
Anomaly detection aims to find instances that are considered unusual and is a fundamental problem of data science. Recently, deep anomaly detection methods were shown to achieve superior results particularly in complex data such as images. Our work focuses on deep one-class classification for anomaly detection which learns a mapping only from the normal samples. However, the non-linear transformation performed by deep learning can potentially find patterns associated with social bias. The challenge with adding fairness to deep anomaly detection is to ensure both making fair and correct anomaly predictions simultaneously. In this paper, we propose a new architecture for the fair anomaly detection approach (\emph{Deep Fair SVDD}) and train it using an adversarial network to de-correlate the relationships between the sensitive attributes and the learned representations. This differs from how fairness is typically added namely as a regularizer or a constraint. Further, we propose two effective fairness measures and empirically demonstrate that existing deep anomaly detection methods are unfair. We show that our proposed approach can remove the unfairness largely with minimal loss on the anomaly detection performance. Lastly, we conduct an in-depth analysis to show the strength and limitations of our proposed model, including parameter analysis, feature visualization, and run-time analysis.
\end{abstract}



\maketitle

\section{Introduction}
Anomalies are the unusual, unexpected, surprising patterns in the observed world that warrant further investigation. Classic work \cite{hawkins1980identification} defines outliers as an observation that deviates so significantly from other observations as to arouse suspicion that a different mechanism generated it. Anomalies and outliers are often used interchangeably though we note that some use the term differently \cite{chandola2009anomaly} and for this paper we use the term anomalies. The goal of an anomaly detection algorithm is given a set of instances to determine which instances stand out as being dissimilar to other instances. Effective detection of anomalies can be used for various applications, such as stopping malicious intruders, fraud detection, system health monitoring, and medical image analysis \cite{chalapathy2019deep}. 

Recent algorithmic developments have proposed many novel deep learning methods for anomaly detection \cite{erfani2016high, chen2017outlier, ruff2018deep, golan2018deep, pang2019deep, hendrycks2019using}. This previous works on deep anomaly detection are typically unsupervised (e.g., assume all training data are from the normal group) and have demonstrated better anomaly detection performance than traditional anomaly detection approaches. One popular approach to deep anomaly detection is the deep support vector data description (deep SVDD) \cite{ruff2018deep}. This work attempts to transform the input data into a new feature space where all the points are closely clustered into a predetermined center. Hence, by definition, those points that cannot be projected to be close to the center are deemed anomalies. The anomaly scores are calculated based on the Euclidean distances between the test instances and the predetermined center during the test time. Deep SVDD is a general approach which can be applied to both low dimensional and high dimensional data. In this first paper on the topic we focus on adding fairness to deep SVDD.

Since anomaly detection is often applied to humans who are then suspected of unusual behavior, ensuring fairness becomes paramount. The notion of fairness has recently received much attention in supervised learning \cite{zafar2017fairness, donini2018empirical} and unsupervised learning \cite{chierichetti2017fair, schmidt2019fair, backurs2019scalable}.  Measures of fairness can generally be divided into two categories \cite{chouldechova2018frontiers}: (i) group-level fairness and (ii) individual level fairness. In anomaly detection problems, we divide the data into two groups, which are the normal group and the abnormal group. We propose to study the group-level fairness problems which ensure that no one particular group contains a disproportionate number of individuals with protected status.  To our best knowledge, there is no prior published work on fairness in the context of deep anomaly detection though work on auditing (i.e., checking) anomaly detection algorithms exist \cite{davidson2020framework}.

\begin{figure}[t]
 \centering
 \vskip -0.1in
 \subfloat[Normal Group]{
 \includegraphics[width=0.23\textwidth]{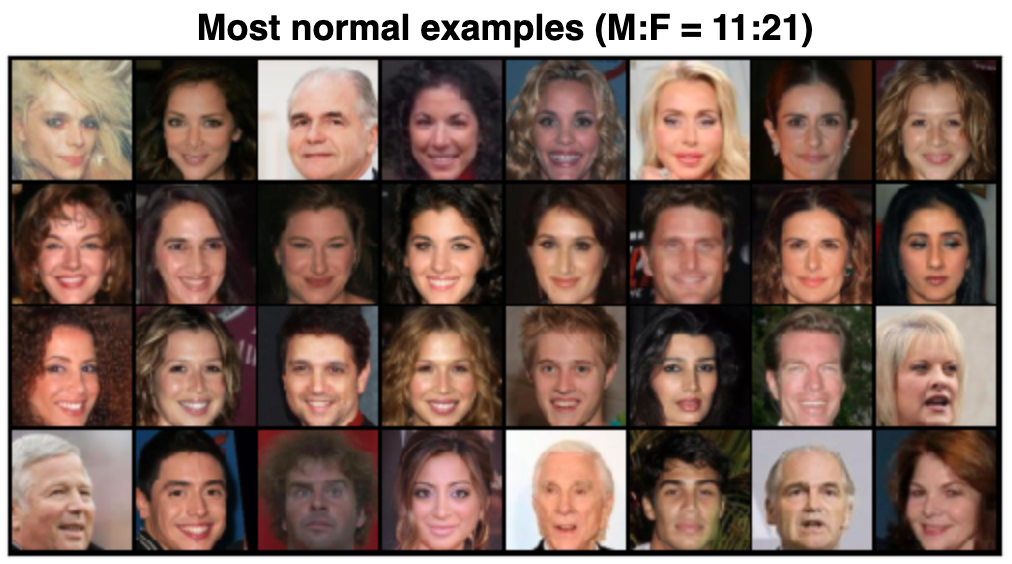}}
 \label{1a}
 \hfill
 \subfloat[Abnormal Group]{
 \includegraphics[width=0.23\textwidth]{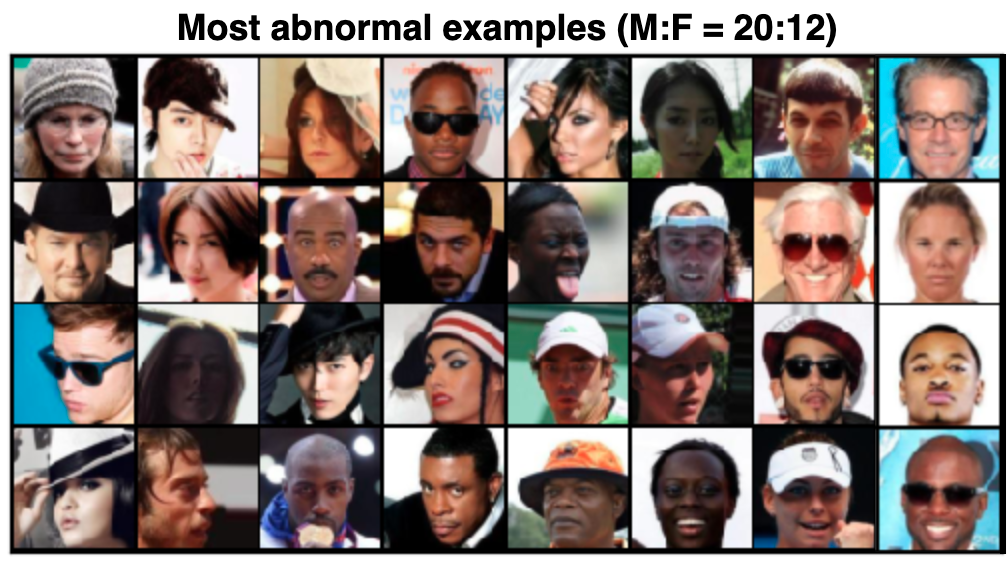}}
 \label{1b}
 \caption{Motivating example of the need for group-level fairness in deep anomaly detection problem. We visualize the top $32$ normal instances and top $32$ abnormal instances discovered by deep SVDD on celebA data set. We see that the normal group is dominated by females while the abnormal group is dominated by males. }
\label{fig:demo}
\vskip -0.15in
\end{figure}

\textbf{A Motivating Example For Group-Level Fairness}. Consider the example of finding anomalies by applying deep SVDD to facial images. The top $32$ normal instances and top $32$ abnormal instances are shown in Figure \ref{fig:demo}. These pictures are from the celebA (celebrity) data set (which we introduce in section  \ref{sec:dataset}). The deep SVDD model is trained on attractive celebrity faces (normal group) and used to detect plain celebrity faces (abnormal group) where the labels are given in the data set. The model performs well in terms of the anomaly detection quality as most attractive celebrity faces and plain celebrity faces are separated correctly. However, when we consider the protected status variable gender in this problem, more females are predicted to be attractive (normal group), and more males are predicted as plain (abnormal group). Moreover, if we consider race as a protected status variable, we can see that the most attractive faces are white people and many black people in the abnormal group. Motivated by these observations, we aim to design experiments to examine the fairness of existing deep anomaly detection methods quantitatively and propose a fair anomaly detection model to balance the number of instances with different sensitive attribute values in the anomaly predictions. 

In this paper, we present the Deep Fair Support Vector Data Description (\emph{Deep Fair SVDD}) model which learns a compact and fair description of the normal data via adversarial learning. We summarize the main contributions in this paper as follows: 
\begin{itemize}
    \item We show existing deep anomaly detection approaches are unfair (see section \ref{sec:unfair_dod}) due to the deep learners' ability to extract out complex features. 
    \item We consider fair anomaly detection in the context of deep representation learning. To the best of our knowledge, this is an under-studied so far and challenging due to the need for fair and high-quality predictions.
    \item We address these challenges by proposing a novel fair anomaly detection architecture (see Figure \ref{fig:pipeline}) and use adversarial learning to remove the unfairness. The idea of using adversarial learning contrasts with many recent works on fairness in learning which typically encodes fairness as a regularization term or a constraint.
    \item We propose two measures of group-level fairness for deep anomaly detection problems: i) A demographic parity motivated fairness measure for the abnormal group (equation \ref{eq:fairness}) ii) A parameter-free measure based on Wasserstein distance for calculating the overall fairness (equation \ref{eq:distribution_distance}).
    \item We demonstrate our method on several types of data, including traditional tabular datasets, face data sets, and digit images. We study the fairness problem concerning gender, racism, and the source of the visual objects. (see section \ref{sec:dataset}).  We find that introducing fairness causes a marginal drop in anomaly detection performance measured by the AUC score (see section \ref{sec:fair_svdd}).
    \item We conduct an in-depth analysis of our proposed model to show our proposed model's strengths and limitations, including parameter analysis, feature visualization, and run-time analysis. (see section \ref{sec:trade_off}, \ref{sec:training_analysis}, \ref{sec:runtime_analysis}).
\end{itemize}

Our paper structure is as follows. In the next section \ref{sec:related_work}, we discuss the related work. Then, we provide background knowledge about deep SVDD and our fairness measures in section \ref{sec:background}. Next, we propose the deep fair SVDD model and analyze how we use adversarial networks to tackle fair anomaly detection problems (section \ref{sec:Method}). Finally, we perform experiments on real-world data sets to demonstrate the effectiveness of our method in section \ref{sec:experiments} and conclude our proposed approach in section \ref{sec:conclusion}.

\section{Related Work}
\label{sec:related_work}

\textbf{Deep Anomaly Detection.} We first outline related works on deep anomaly detection. One of the most common deep anomaly detection approaches is reconstruction-based methods \cite{hawkins2002outlier, masci2011stacked, xia2015learning, an2015variational, sakurada2014anomaly, chen2017outlier, huang2019inverse} which assume the anomalies possess different features than the normal instances. Hence, given a pre-trained autoencoder over the normal instances it will be hard to compress and reconstruct the anomalies. The anomaly score in this research is defined as the reconstruction loss for each test instance. Inspired by the generative adversarial networks \cite{goodfellow2014generative}, another line of related works \cite{schlegl2017unsupervised, deecke2018image, zenati2018efficient} score an unseen sample based on the ability of the model to generate a similar one. 

More recently, A deep version of support vector data description (Deep SVDD) has been proposed \cite{ruff2018deep}. This work is inspired by kernel-based one-class classification \cite{scholkopf2001estimating} which combines the ability of deep representation learning with the one-class objective to separate normal data from anomalies by concentrating normal data in embedded space while mapping anomalies to distant locations. Another recent progress on deep anomaly detection uses self-supervised learning on image data sets and achieves excellent performance \cite{gidaris2018unsupervised, golan2018deep, hendrycks2019using, wang2019effective}. For example, \cite{golan2018deep} uses a composition of image transformations and then trains a neural network to predict which transformation was used. The anomaly scores are computed based on the predictions' confidence over different image transformations given the test samples. 


\textbf{Fairness in Anomaly Detection.} With so many works focusing on improving the deep anomaly detection performance, our work differentiates from those previous works as we investigate the fairness of the existing deep anomaly detection problems and propose a novel deep fair anomaly detection model to help humans make fair decisions. To the best of our knowledge, there is no work on deep fair anomaly detection algorithms. We now introduce two related works on non-deep fair anomaly detection problems. Recent work \cite{davidson2020framework} has studied auditing the output of any anomaly detection algorithm. In their work, the anomaly detection algorithm’s output fairness with respect to multiple protected status variables (PSVs) is measured by finding PSV combinations in the outlier group which are more common than in the normal group. Their empirical results show that the output of five classic anomaly detection methods is unfair. Another work \cite{deepak2020fair} studies the fairness problem of LOF (Local Outlier Factor) \cite{breunig2000lof} and proposes several heuristics to mitigate the unfairness within LOF on tabular data sets. Differently, our work proposes to examine fairness for the deep anomaly detection problems which work for both tabular data and image data. Moreover, unlike LOF-based approaches that have no training phase and do not learn a model of normality, our proposed model can make out-of-sample predictions. 

\textbf{Adversarial Learning for Fairness.} Lastly, we introduce the related works which take the advantages of adversarial networks to remove unfairness. \cite{beutel2017data} applies an adversarial training method to satisfy parity for salary prediction. This work shows that small amounts of data are needed to train a powerful adversarial model to enforce fairness constraints. The work of \cite{zhang2018mitigating} uses a predictor and adversary with an additional projection term to remove unfairness in both supervised learning tasks and debiasing word embedding tasks. \cite{elazar2018adversarial} shows that demographic information leaks into intermediate representations of neural networks trained on text datasets and applies adversarial learning to mitigate the information leaks. \cite{sweeney2020reducing} takes the advantages of adversarial networks to reduce word vector sentiment bias for demographic identity terms.

\section{Preliminary}
\label{sec:background}
\subsection{Deep Support Vector Data Description}
Among the recent deep anomaly detection methods we focus on deep SVDD \cite{ruff2018deep} as a base learner because it is not only a popular method but also performs well on both low dimensional (tabular data) and high dimensional data (images). 
Unlike generative models or compression based anomaly detection models which are adapted for anomaly detection, deep SVDD is directly learned with an anomaly detection based objective. 
Given the training data of just normal points $\X \in {\R}^{n \times d}$, the deep SVDD network is trained to map all the $n$ normal points close to a fixed center $\c$ where $\c$ is normally set as the mean of the  points. Denote function $f$ as a neural network with parameters $\theta$ the simplified objective function is:
\begin{equation}
    \argmin_{\theta} \frac{1}{n} \sum_{i=1}^{n}{||f({\x}_i;\theta) - \c||}^2 + \frac{\alpha}{2} \sum_{\ell = 1}^{L} {||{\theta}^{\ell}||}^2
\label{deepsvdd}
\end{equation}
The second term is a network weight decay regularizer with hyper-parameter $\alpha > 0$ which prevents finding a too complex mapping function. The network has $L$ hidden layers and set of weights $\{{\theta}^1, ... , {\theta}^L \}$ are the weights of layer $\ell \in \{1, ..., L\}$. Deep SVDD contracts the embedding space enclosing the points by minimizing the mean distance of all data points to the center. During the evaluation/scoring stage, given a test point $\x \in {\X}^T$ Deep SVDD will calculate the anomaly score $s(\x)$ for $\x$ as follows: 
\begin{equation}
    s(\x) = {||f(\x;\theta) - \c||}^2
\end{equation}
Note this is just the distance the instance is from the center, abnormal points are then those far from the center.

\subsection{Notion of Fairness}
Fairness is measured using protected status variables or sensitive features such as gender and race. In this paper, we study group-level fairness which ensures that no one particular group contains a disproportionate number of instances of a given protected status. 

\textbf{Fairness by $p \%$ -rule.} Our first notion of fairness is inspired by \cite{zafar2017fairness} which proposed a statistical parity motivated measure for a supervised classification model. Statistical parity is a popular fairness measure used in many unsupervised learning and supervised learning problems \cite{chierichetti2017fair, backurs2019scalable, zafar2017fairness, slack2020fairness}. Let $t$ be the anomaly score threshold, then the normal groups are points with $s(\x) \le t$ and the abnormal groups are points with $s(\x) > t$. Given the protected status variable as $z \in \{0, 1\}$, our definition of fairness measure leverages the $80\%$ rule \cite{biddle2006adverse}: a normal / abnormal group partition satisfies the $80\%$ rule if the ratio between the percentage of person with a particular protected status variable value having $s(\x) > t$ and the percentage of person without protected status having $s(\x) > t$ is no less than $80:100$. We define the $p \%$ -rule as our fairness measure for the anomaly detection problem:

\begin{equation}
     \min (\frac{P(s(x) > t | z=1)}{P(s(x) > t | z=0)}, \frac{P(s(x) > t | z=0)}{P(s(x) > t | z = 1)}) \ge \frac{p}{100}
    \label{eq:fairness}
\end{equation}
Note the $p \%$ -rule value ranges from $0$ to $1$ and a larger value indicates the model is fairer. In ideal case we have $P(s(x) > t | z=1) = P(s(x) > t | z=0)$. Maximizing $p \%$ -rule means predicting $x$ as an anomaly will be independent of the protected status variable $z$

The rationale behind using our first fairness measure in equation \ref{eq:fairness} is because it is closely related to the $80\%$ rule advocated by the US Equal Employment Opportunity Commission \cite{biddle2006adverse}. We can determine a deep anomaly detection model's fairness using the $80\%$ rule. However, there are some limitations to our first proposed measurement. Firstly, we need to know the exact number of anomalies in the test set to correctly set the anomaly score threshold $t$ to partition the normal and abnormal groups. Secondly, this measure only considers the fairness in the abnormal group. 

 \begin{figure*}[th]
 \vskip -0.15in
 \centering
 \subfloat[Model \emph{A}'s Predictions]{
 \includegraphics[width=0.24\textwidth]{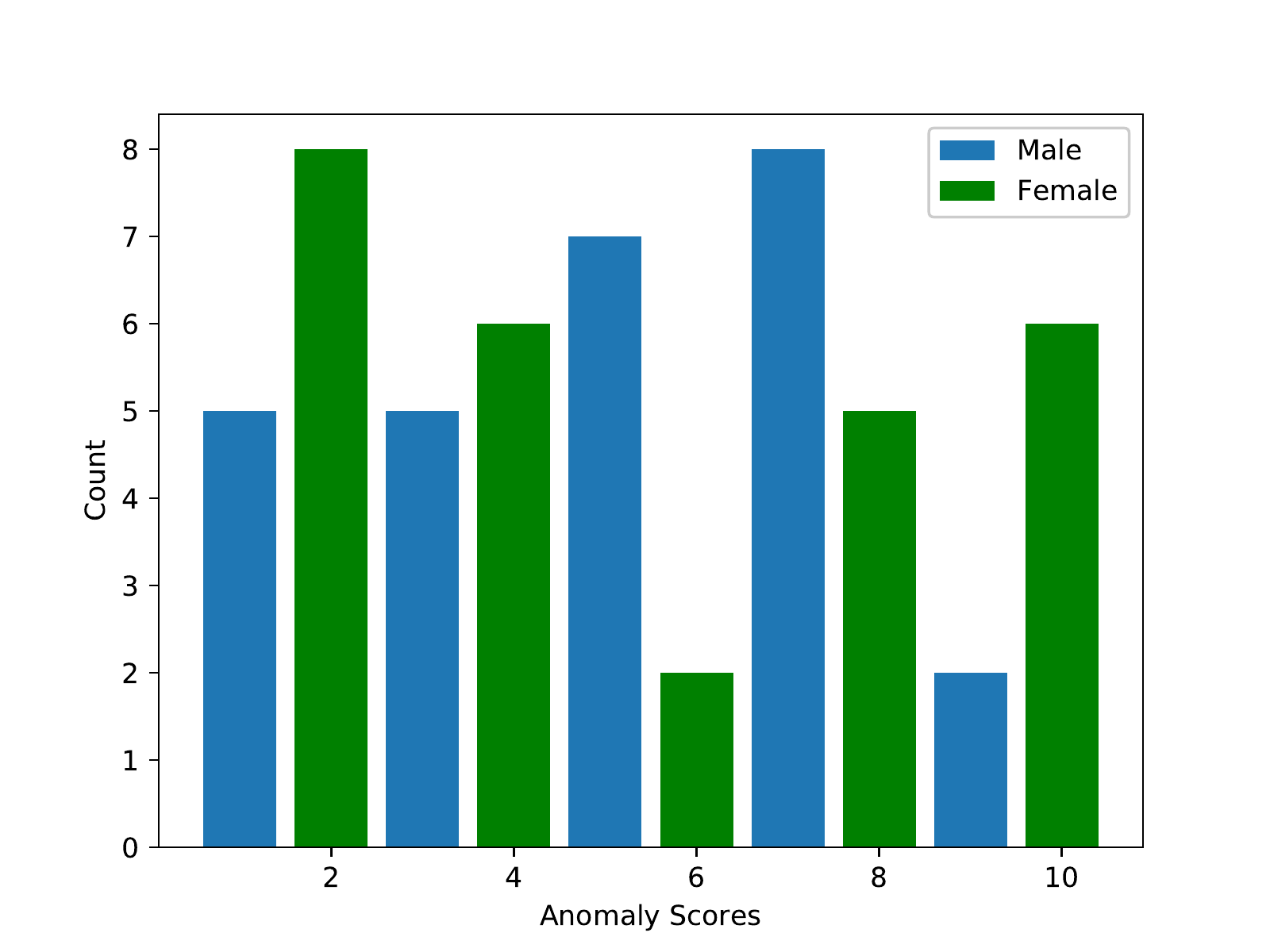}}
 \label{2a}
 \hfill
 \subfloat[Model \emph{B}'s Predictions]{
 \includegraphics[width=0.24\textwidth]{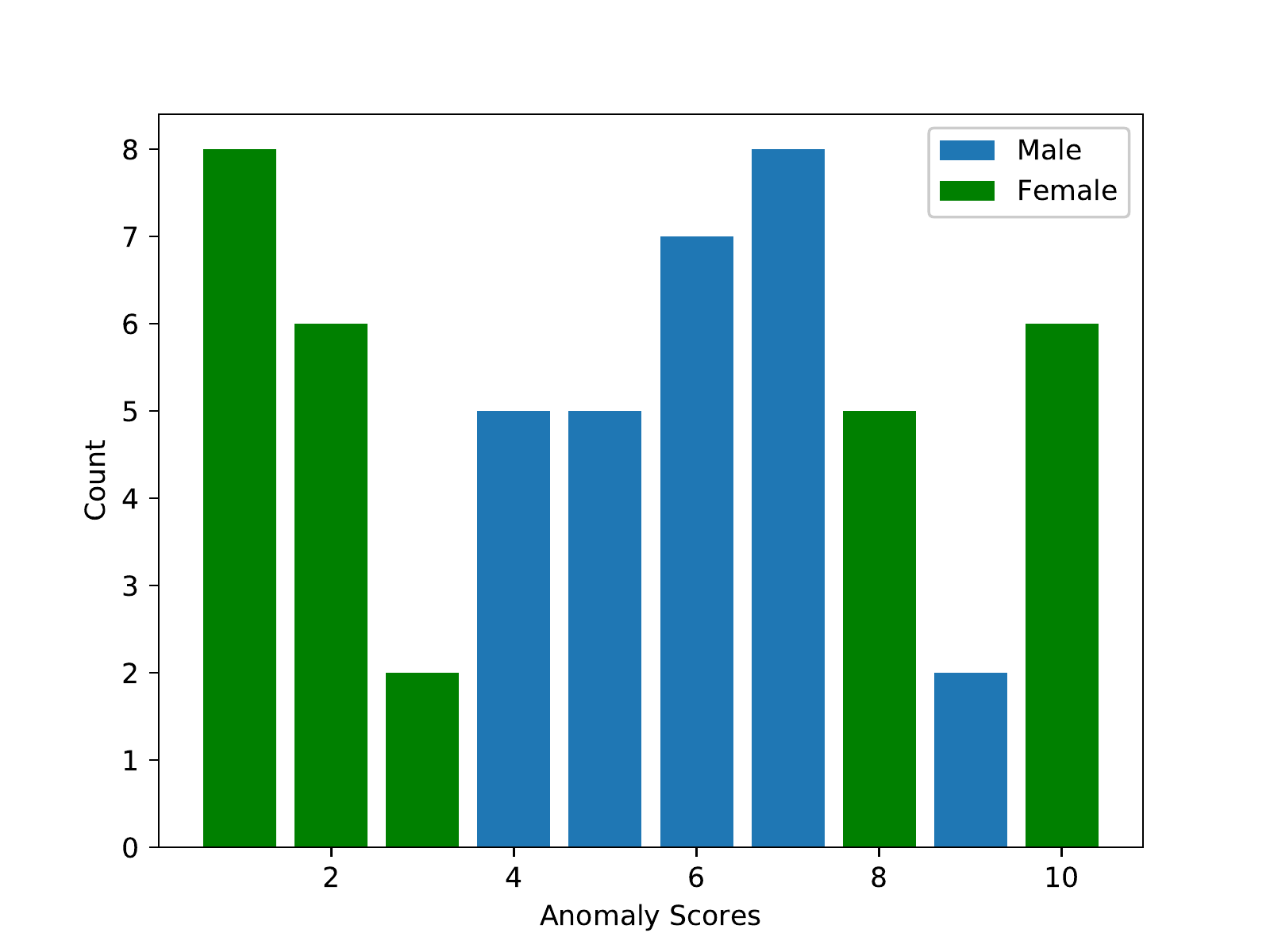}}
 \label{2b}
 \hfill
  \subfloat[Model \emph{A}'s Prediction Distribution]{
 \includegraphics[width=0.24\textwidth]{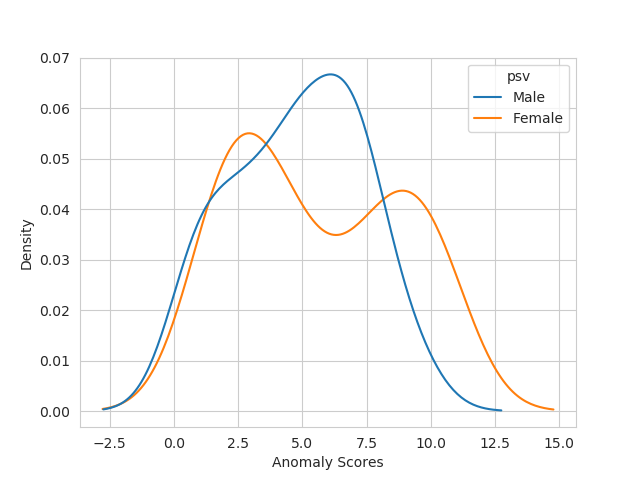}}
 \label{2c}
 \hfill
 \subfloat[Model \emph{B}'s Prediction Distribution]{
 \includegraphics[width=0.24\textwidth]{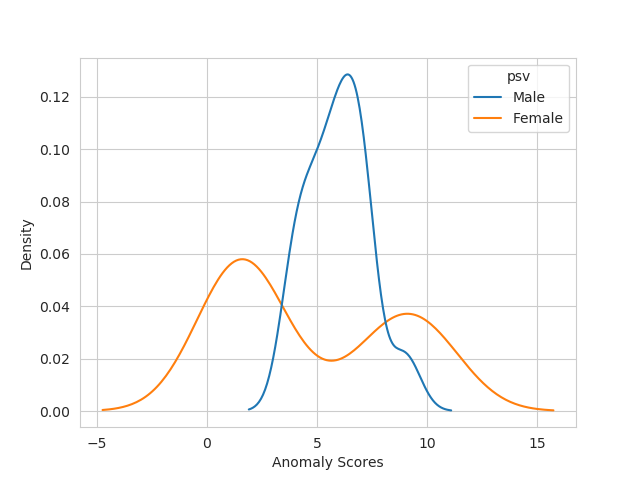}}
 \label{2d}
\caption{A toy example to show the difference between our proposed two fairness measures. Figure a, b summaries the statistics of predicted anomaly scores of model \emph{A} and \emph{B}. Given the ground-truth anomaly score threshold $t = 8$,  model \emph{A} and \emph{B} have the same fairness by $p \%$ -rule as $2 / 6 = 0.33$. Figure c, d shows the anomaly score distributions for model \emph{A}'s predictions ($M_A, F_A$) and model \emph{B}'s predictions ($M_B, F_B$). Model \emph{B} is more unfair as the anomaly scores are highly correlated with the sensitive attribute \emph{gender} ($M, F$). The fairness by distribution distance for model \emph{A} and \emph{B} are $W(M_A, F_A) = 1.37$ and $W(M_B, F_B) = 2.87$.}
\label{fig:metric_demo}
\end{figure*}
\textbf{Fairness by distribution distance.} Here we propose a new fairness measure for anomaly detection problems which is invariant of the anomaly score threshold $t$ and covers both normal and abnormal groups.
We have designed one synthetic anomaly detection problem to show the motivation for our second fairness measure. Assume there are two anomaly detection models named \emph{A} and \emph{B}. The test data includes $27$ males and $27$ females, and the binary sensitive attribute is $gender$. To be specific, the predicted anomaly scores from Model \emph{A} and \emph{B} are shown in Figure \ref{fig:metric_demo} (a) and (b). Given the ground truth number of anomalies as $8$, we can set the anomaly score threshold $t = 8$ to predict anomalies with $s(x) > 8$. Now we can calculate the $p \%$ -rule for Model \emph{A} and Model \emph{B} as: $ 2 / 6 = \frac{1}{3}$. Although models \emph{A} and \emph{B} achieve the same fairness measured by $p \%$ -rule, we can learn from the anomaly score distributions in Figure \ref{fig:metric_demo} (c) and (d) that model \emph{B}'s predictions are highly correlated with the sensitive attribute \emph{gender} which is less fair.  


Now we formulate our second definition of fairness which quantifies the difference between each demographic group's anomaly score distributions: let $\mathbb{P}$ denotes the distribution of the anomaly scores for test instances with sensitive attribute $z = 0$ and $\mathbb{Q}$ for test instances with sensitive attribute $z = 1$. We calculate the Wasserstein-1 (\emph{Earth-Mover Distance}) distance between distribution $\mathbb{P}$ and $\mathbb{Q}$ as fairness by distribution distance measure:
\begin{equation}
  W(\mathbb{P}, \mathbb{Q}) = \inf_{\gamma \in \Pi (\mathbb{P}, \mathbb{Q})} {\E}_{(x, y) \sim \gamma} [||x - y||]
    \label{eq:distribution_distance}
\end{equation}
where $\Pi(\P, \mathbb{Q})$ denotes the set of all joint distributions $\gamma(x, y)$ whose marginals are respectively $\P$ and $\mathbb{Q}$. Intuitively, $\gamma(x, y)$ indicates how much "mass" must be transported from $x$ to $y$ in order to transform the distribution $\P$ to $\mathbb{Q}$. For our previous toy example, we calculate the \emph{Distribution distance} for model \emph{A} and \emph{B}'s predictions as $1.37$ and $2.78$. These results indicate that model \emph{A} is overall fairer than model \emph{B}. From a practitioner's perspective, we can use distribution distance to evaluate the fairness performance for different anomaly detection models and conduct model selection when we have no access to the test set. Lastly, we will use both \emph{Fairness by $p \%$ -rule} and \emph{Fairness by distribution distance} measures to evaluate the fairness performance in our experimental section.


\section{Methods}
\label{sec:Method}
\subsection{Learning Overview}
In this section, we propose the deep fair SVDD model for deep anomaly detection problems. Following the previous deep anomaly detection works \cite{ruff2018deep, gidaris2018unsupervised}, we assume the training data $\X$ contains only normal instances. Moreover, our proposed model requires access to the binary protected status variable $Z$ for each of the training instances $\X$. We learn $f(\theta)$ as an encoder network to learn compact descriptions of $\X$ (i.e. a mapping to a lower-dimensional space), and a classification network $g(\theta_{d})$ to predict 
protected status variable value $z \in Z$ based on the learned embedding $f(\X; \theta)$. 
We train the encoder $f$ and discriminator $g$ using adversarial training so that we hope the embedding learned via encoder $f$ can fool the discriminator $g$. 
Training such a network is challenging and we take advantage of adversarial learning since it has shown promising results for other fairness tasks such as removing unfairness in NLP applications \cite{elazar2018adversarial,sweeney2020reducing}. We use adversarial learning to de-correlate the relationships between protected status variable $Z$ and feature vectors encoded via $f(\theta)$. Note that our fair learning method is fundamentally different from much existing work \cite{zafar2017fairness,celis2019classification,hu2020fair} which uses a regularization term to encode fairness or encodes fairness as a constraint.



 \begin{figure*}[th]
\centering
\includegraphics[width=\textwidth]{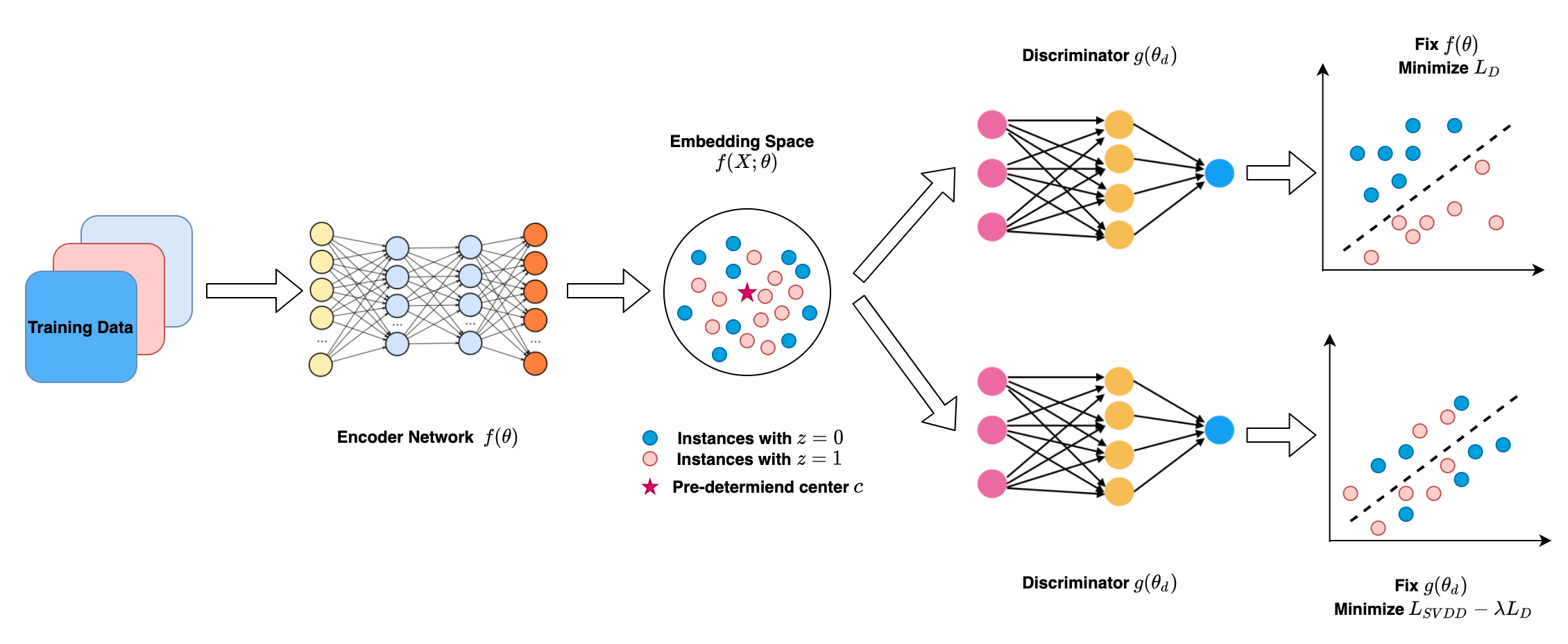}
\caption{ Pipeline of the proposed deep fair SVDD learning framework. The inputs are normal training data $\X$ and the outputs are learned embedding $f(\X;\theta)$ and a discriminatory function $g(\theta_d)$. The end-to-end learning process contains three steps: 1) train the encoder $f(\theta)$  via minimizing the loss ${L}_{SVDD}$, 2) fix the encoder's parameters $\theta$,  and train the discriminator $g(\theta_{d})$ via minimizing the discriminator's loss ${L}_D$, 3) fix the discriminator's parameters ${\theta}_d$ and train encoder $f(\theta)$ to minimize the adversarial loss $L_{SVDD} - \lambda L_D$. Procedure (2) and (3) are trained alternatively until convergence.}
\label{fig:pipeline}
\end{figure*}

\subsection{Deep Fair SVDD Model}
 Our proposed deep fair SVDD network aims to learn a fair representation to describe all the training data via adversarial learning. Given the normal training data $\X \in {\R}^{M \times D}$, encoder network $f(\theta)$ we have the latent encoding of all the normal points as $f(\X; \theta)$. Assume the binary protected status variable is $Z \in {\R}^{M \times 1}$. The fair representation is achieved when the learned embedding are statistically independent of sensitive attribute $Z$. Given $z \in \{0, 1\}$ we hope to optimize the function $f(\theta)$ to have:
 \begin{equation}
     \p (f(\X; \theta) | z = 0 ) = \p (f(\X; \theta) | z = 1)
     \label{fair_goal}
 \end{equation}
 To achieve the goal in equation (\ref{fair_goal}) we propose to use adversarial networks with a min-max game strategy to constrain the embedding function $f(\theta)$.
 Firstly, the encoder network is trained with normal points $\X$ to generate compact embedding around a pre-determined center $\c$. To regularize the encoder we add a weight decay regularizer with positive hyper-parameter $\alpha$ for all the $L$ hidden layers. We use term $L_{SVDD}$ to represent the encoder's loss function:
 \begin{equation}
     L_{SVDD} = \frac{1}{M} \sum_{i=1}^{M}{||f({\x}_i;\theta) - \c||}^2 + \frac{\alpha}{2} \sum_{\ell = 1}^{L} {||{\theta}^{\ell}||}^2
     \label{svdd_loss}
 \end{equation}
 Secondly we concatenate the encoding network $f(\theta)$ with a discriminator $g(\theta_d)$ to learn to classify the sensitive attributes $Z$ based on learned embedding $f(\X; \theta)$. Since $Z$ is a binary variable we use sigmoid function to get the probabilistic prediction as $\hat{z}_i$:
 \begin{equation}
     \hat{z}_i = \frac{1}{1 + {\exp}^{-g(f({\x}_i;\theta) | \theta_d)}}
 \end{equation}
 We choose cross entropy loss to train discriminator $g(\theta_d)$ as:
 \begin{equation}
     L_{D} = -\frac{1}{M} \sum_{i=1}^{M} ( z_i * \log(\hat{z}_i) + (1-z_i)*  \log(1 - \hat{z}_i))
     \label{discriminator_loss}
 \end{equation}
 To make the learned embedding $f(\X; \theta)$ invariant with sensitive attributes $Z$ we hope to tune the embedding function $f(\theta)$ to fool the discriminator $g(\theta_d)$. Meanwhile, we hope the normal points are still closely clustered together so that we design the adversarial loss $L_{Adv}$ as follows:
 \begin{equation}
     L_{Adv} = L_{SVDD} - \lambda L_{D}
     \label{adv_loss}
 \end{equation}
 where the hyper-parameter $\lambda$ is a positive constant number. Minimizing the adversarial loss $L_{Adv} = L_{SVDD} - \lambda L_{D}$ is actually maximizing the discriminator's loss $L_{D}$. Note the discriminator's parameters $\theta_d$ are fixed when we back-propagate the adversarial loss. Similar as the generated adversarial networks \cite{goodfellow2014generative}, we propose to train the $f(\theta)$ and $g(\theta_d)$ in an alternative way until we find the min-max solution. The training procedure tries to jointly optimize both quantities:
 \begin{equation}
     \argmin_{\theta_{d}} L_{D}
 \end{equation}
 \begin{equation}
     \argmin_{\theta} L_{SVDD} - \lambda L_{D}
 \end{equation}

 Once the joint training converges, the anomaly scores for all the instances are calculated as:
 \begin{equation}
  {\S} = {||f(\X;\theta) - \c||}^2
  \label{eq:anomaly_scores}
 \end{equation}
 Note the instances with larger anomaly scores have larger probability to be predicted as anomalies. The pseudo-code for the learning algorithm is summarized in Algorithm \ref{alg:2}. We also visualize the learning pipeline of deep fair SVDD model in Figure \ref{fig:pipeline}.
\begin{algorithm}[h]
\vskip 0.05in
 \caption{Algorithm for deep fair SVDD}
 \label{alg:2}
\begin{algorithmic}[1]
 \STATE {\bfseries Input:} $\X$: training data, ${\X}^T$: test data, $Z$: Protected status variable, $f(\theta)$: encoder network, $\c$: pre-determined data center, $g(\theta_d)$: discriminator, $K$: initial training epochs, $T$: adversarial training epochs.
 \STATE {\bfseries Output:} ${\S}$: predicted anomaly score.
 \smallskip
 \STATE Train the encoder network $f(\theta)$ via minimizing $L_{SVDD}$ in equation (\ref{svdd_loss}) for $K$ epochs.
 \STATE Fix the encoder network $f(\theta)$, train the discriminator $g(\theta_{d})$ via minimizing $L_D$ in equation (\ref{discriminator_loss}) for $K$ epochs.
 \FOR{epoch from $1$ to $T$}
 \STATE Fix the parameters $\theta$ for encoder network $f(\theta)$. Calculate $L_D$ in equation (\ref{discriminator_loss}) for each mini-batch.
 \STATE Back-propagate the discriminator loss $L_D$ and update the parameters $\theta_d$.
 \STATE Fix the parameters $\theta_d$ for discriminator $g(\theta_{d})$. Calculate the loss $L_{Adv}$ in equation (\ref{adv_loss})  for each mini-batch.
 \STATE Back-propagate the adversarial loss $L_{Adv}$ and update $\theta$.
 \ENDFOR
 \STATE Output the anomaly scores for test set ${\S} = {||f({\X}^T;\theta) - \c||}^2$.
\end{algorithmic}
\end{algorithm}

\begin{table*}
  \caption{ Characteristics of four datasets used in our experiments. Our methods requires the protected status variables such as Gender (Male and Female) and Race (African-American and non African-American) to be binary variables.}
  \label{tab:datasets}
  \begin{tabular}{c|c|c|c|c|c|c}
    \toprule
    Dataset &Type & $\#$ Instances & $\#$ Dimension &Protected Status Variable & Normal Group & Abnormal Group\\
    \midrule
    COMPAS Recidivism \cite{angwin2016machine} & Tabular   & 3878    &11            & Race  & Not reoffending& Reoffending\\
    \hline
    celebA \cite{liu2015faceattributes}  & Face      & 24000   &64 x 64 x 3  & Gender   & Attractive faces & Plain faces\\
    \hline
    MNIST-USPS      & Digits    & 7435    &28 x 28 x 1  & Source of digits        & Digit 3 & Digit 5\\
    \hline
    MNIST-Invert    & Digits    & 15804   &28 x 28 x 1  & Color of the digits     & Digit 3 & Digit 5\\
  \bottomrule
\end{tabular}
\end{table*}
\subsection{Potential Extensions of Deep Fair SVDD}
\label{sec:variations}
In this subsection, we analyze the design of our proposed deep fair SVDD and provide several potential extensions of our proposed learning framework that we intend to study: 
    \subsubsection{Extensions to Fairness Problems with Multi-State Protected Status Variable}
    Note we study the fairness problem with binary protected status variable $z \in \{0, 1\}$ in this work. However, our deep fair SVDD learning framework can be extended to solve fairness problems with multi-state protected state variable (e.g., education level, nationality) by changing the current binary discriminator $g$ into a multi-class classification network.
    
    \subsubsection{Extensions to Fairness Problems with Multiple Protected Status Variable}
    Our framework can also support multiple protected status variables if we substitute the binary classification discriminator with a multi-class classification network. Given the fairness requirements on multiple protected state variables (say gender and race together), we can enumerate all the combinations via a Cartesian product of these two variables and transform them into a multi-state protected state variable to feed in our extended framework. This is an important property lacking in many fair classification methods as clearly making a model fairer with respect to say gender could make it unfair with respect to say race. 
    
    \subsubsection{Extensions to Semi-supervised Anomaly Detection}
    The current encoder $f(\theta)$ is trained via an unsupervised loss function ($\ref{svdd_loss}$) to force all the normal data to be close to the pre-determined center $c$. Recently, some works on general semi-supervised anomaly detection \cite{gornitz2013toward,ruff2019deep} have demonstrated superior performance. In general semi-supervised anomaly detection settings, we assume the learners have access to a small subset of labeled normal and abnormal instances. Our current learning framework can be modified to accommodate semi-supervised anomaly detection settings by combining the current loss function (\ref{svdd_loss}) with a new supervised classification loss for labeled anomalies in the training set.

\section{Experiments}
\label{sec:experiments}
In this section, we conduct experiments to empirically evaluate our proposed approach. From our  experiments, we aim to address following questions:
\begin{itemize}
    \item Do  existing deep anomaly detection algorithms produce unfair results? (see Section \ref{sec:unfair_dod})
    \item How does our proposed algorithms work in two types of datasets: involving low dimensional data (COMPAS Recidivism) and high dimensional data (Facial images and digits)? (see Section \ref{sec:fair_svdd})
    \item What is the sensitivity of the hyper-parameter $\lambda$ in our proposed deep fair SVDD model (see Section \ref{sec:trade_off})?
    \item How do our proposed algorithm change the latent feature space whilst making anomaly detection fairer? (see Section \ref{sec:outlier_predictions}, \ref{sec:training_analysis})
    \item How efficient are our proposed algorithms? (see Section \ref{sec:runtime_analysis})
\end{itemize}

\subsection{Data Sets}
\label{sec:dataset}
We propose to experiment on four public datasets which include visual data and tabular data. We list the characteristics of our selected datasets in Table \ref{tab:datasets} and introduce the details of how we construct those datasets as below. For each data set, only normal instances are in the training data set, but there are both normal and abnormal instances in the test data.
\begin{itemize}
    \item COMPAS Recidivism \cite{angwin2016machine}: The COMPAS recidivism data set consists of data from criminal defendants from Broward County, Florida. We create a binary protected status variable for whether the defendant is African American. Given the ProPublica collected label of whether the defendant was rearrested within two years, we set the normal group for people who were not reoffending and the abnormal group for reoffending. We select this data set to demonstrate our approach's performance on low-dimension tabular data.
    \item celebA \cite{liu2015faceattributes}: This is a large-scale face attributes dataset with more than $200$K celebrity images, each with $40$ attribute annotations. We sample a subset of this data set and treat gender as a binary protected status variable in this data set. The normal group contains celebrity faces labeled as attractive, and the abnormal group contains the celebrity faces labeled as plain. We choose celebA data set to test our approach on high-dimension images.
    \item MNIST-USPS: This dataset consists of MNIST and USPS images, which include different style's hand-written digits. We set the sample source as a binary protected attribute. The normal group contains digits from class $3$, and the abnormal group includes digits from class $5$. 
    \item MNIST-Invert: We take the images from MNIST and create a duplicate which is inverted to build this dataset. The binary protected attribute is then original or inverted. The normal group contains digits from class $3$ and the abnormal group contains digits from class $5$.
\end{itemize}

\begin{table*}[th]
  \caption{ Characteristics of original training set and balanced training set used in experiments. We reduce the number of over-represented group in original training set to generate balanced training set. }
  \label{tab:training_set}
  \begin{tabular}{c|c|c|c|c}
    \toprule
         &COMPAS Recidivism & celebA & MNIST-Invert &MNIST-USPS \\
    \midrule
    Original ($z = 0$)  & 1480    & 16000   &   6000  & 6131  \\
    \hline
    Original ($z = 1$)  & 1210    & 4000    &   6000  & 658  \\
    \hline
    Balanced ($z = 0$)  & 1210    & 4000    &   6000  & 658  \\
    \hline
    Balanced ($z = 1$)  & 1210    & 4000    &   6000  & 658  \\
  \bottomrule
\end{tabular}
\vskip -0.15in
\end{table*}

\subsection{Implementation}
Due to the different characteristics of our selected data sets, we have implemented different networks for them. For the SVDD based encoder network $f(\theta)$: we use a convolutional neural network with two modules, $8$ $(5 \times 5)$ filters followed by $4$ $(5\times5)$ filters, and a final fully connected layer of $32$ units for MNIST-USPS and MNIST-Invert data sets; we use a convolutional neural network with three modules, $32$ $(5\times5\times3)$ filters, $64$ $(5\times5\times3)$ filters, and $128$ $(5\times5\times3)$ filters, followed by a fully connected layer of $128$ units for celebA data set; we use a fully connected neural network with two hidden layers with $32$ and $16$ units for the COMPAS Recidivism data set. We use batch normalization \cite{ioffe2015batch} and ReLU activations in these networks. Note for the fair deep SVDD model we have another classification branch $g(\theta_{d})$. We employ a fully connected neural network with three hidden layers $(500-2000-500)$ as the sensitive attribute discriminator for all the data sets. We set the trade-off hyper-parameter $\lambda$ default to $1$ and the center $c$ as the mean of all the instances embeddings. We set the learning rate as ${1e}^{-3}$ for Adam optimizer and conduct mini-batch training with batch size as $128$. The weight decay hyper-parameter $\alpha$ is set to $5 * {10}^{-6}$.

 \begin{figure*}[h]
 \centering
 \subfloat[Deep SVDD ($p \%$ -rule)]{
 \includegraphics[width=0.24\textwidth]{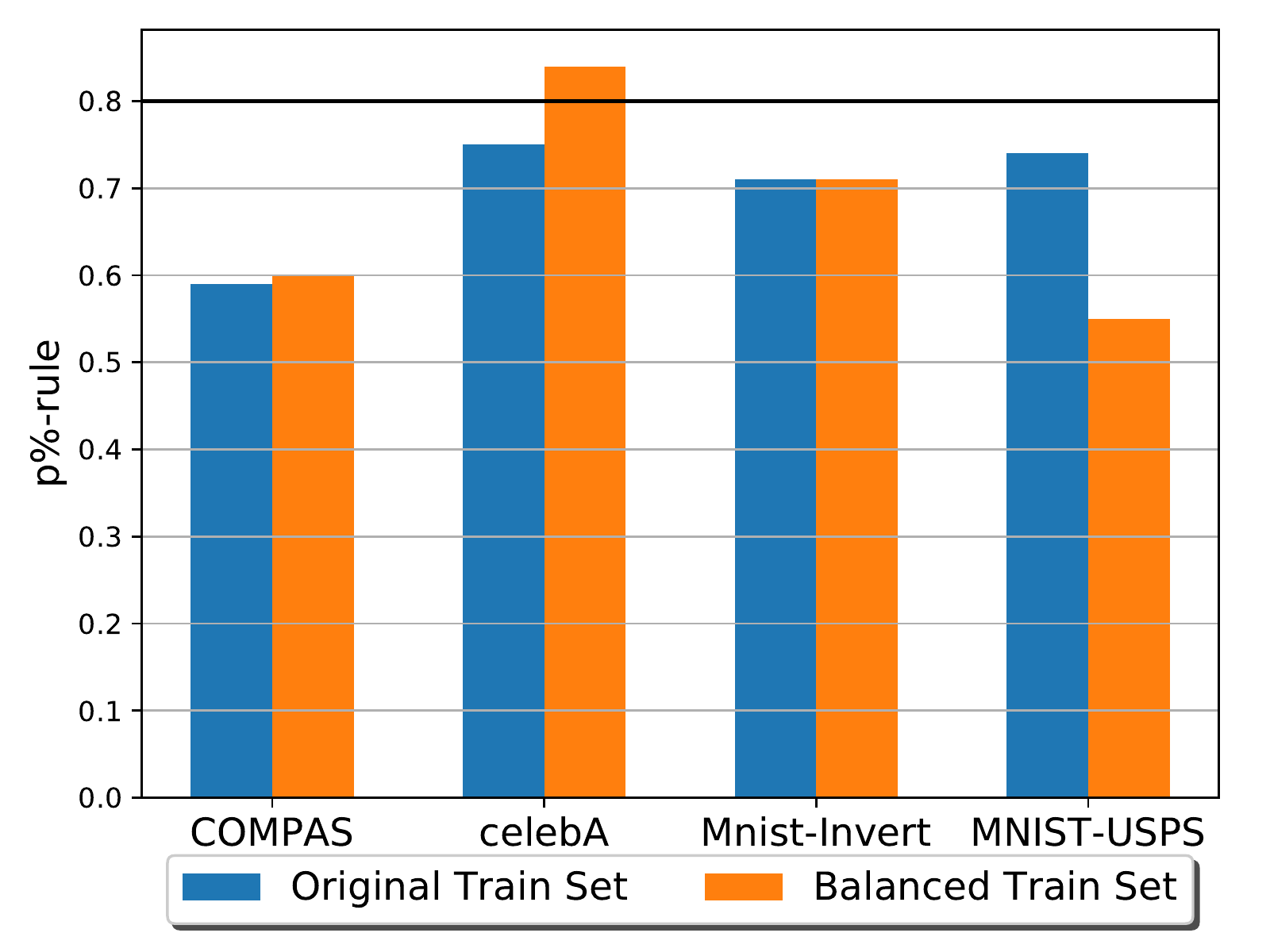}}
 \label{3a}
 \hfill
 \subfloat[DCAE ( $p \%$ -rule)]{
 \includegraphics[width=0.24\textwidth]{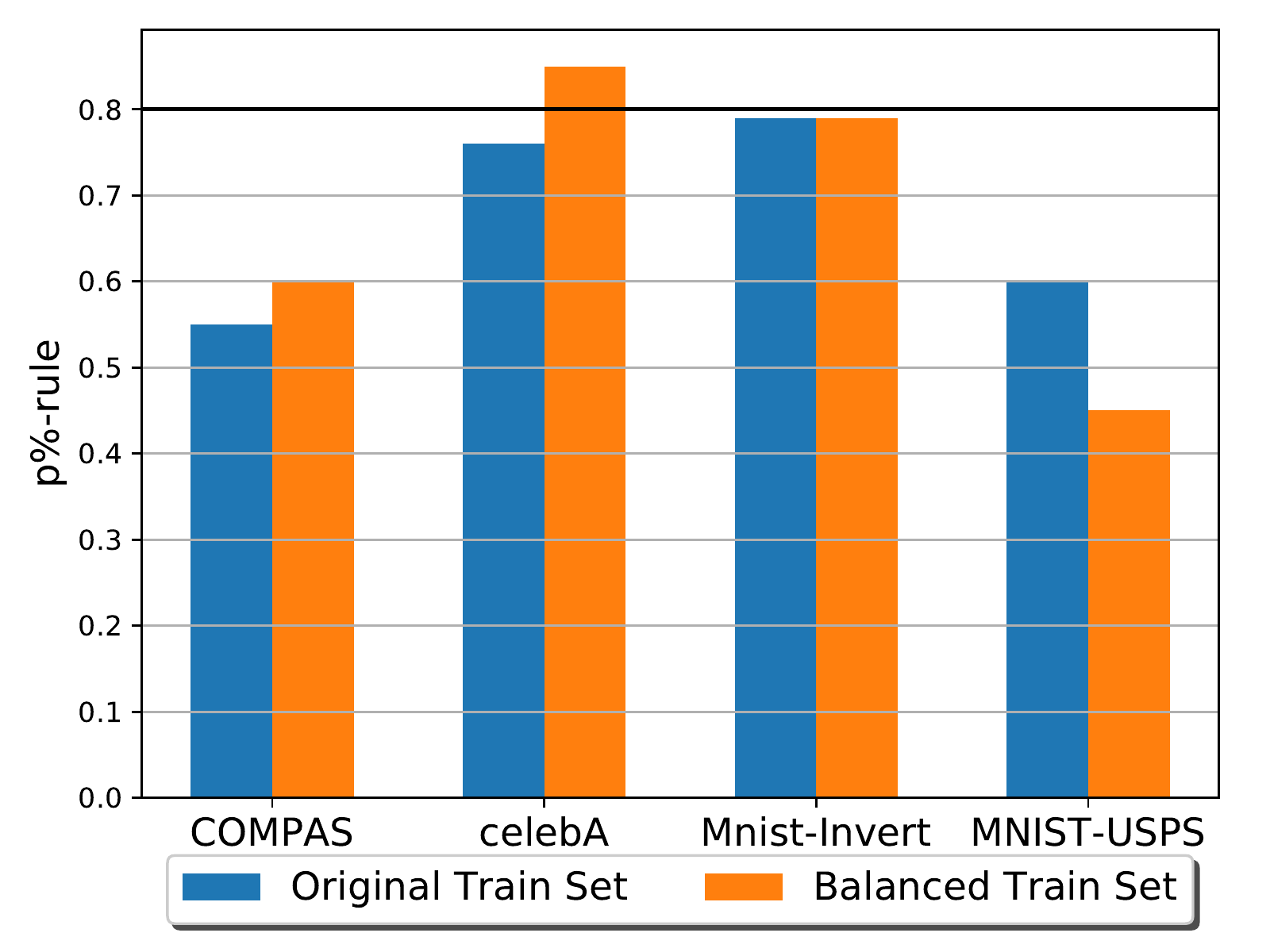}}
 \label{3b}
  \hfill
 \subfloat[Deep SVDD (distribution distance)]{
 \includegraphics[width=0.24\textwidth]{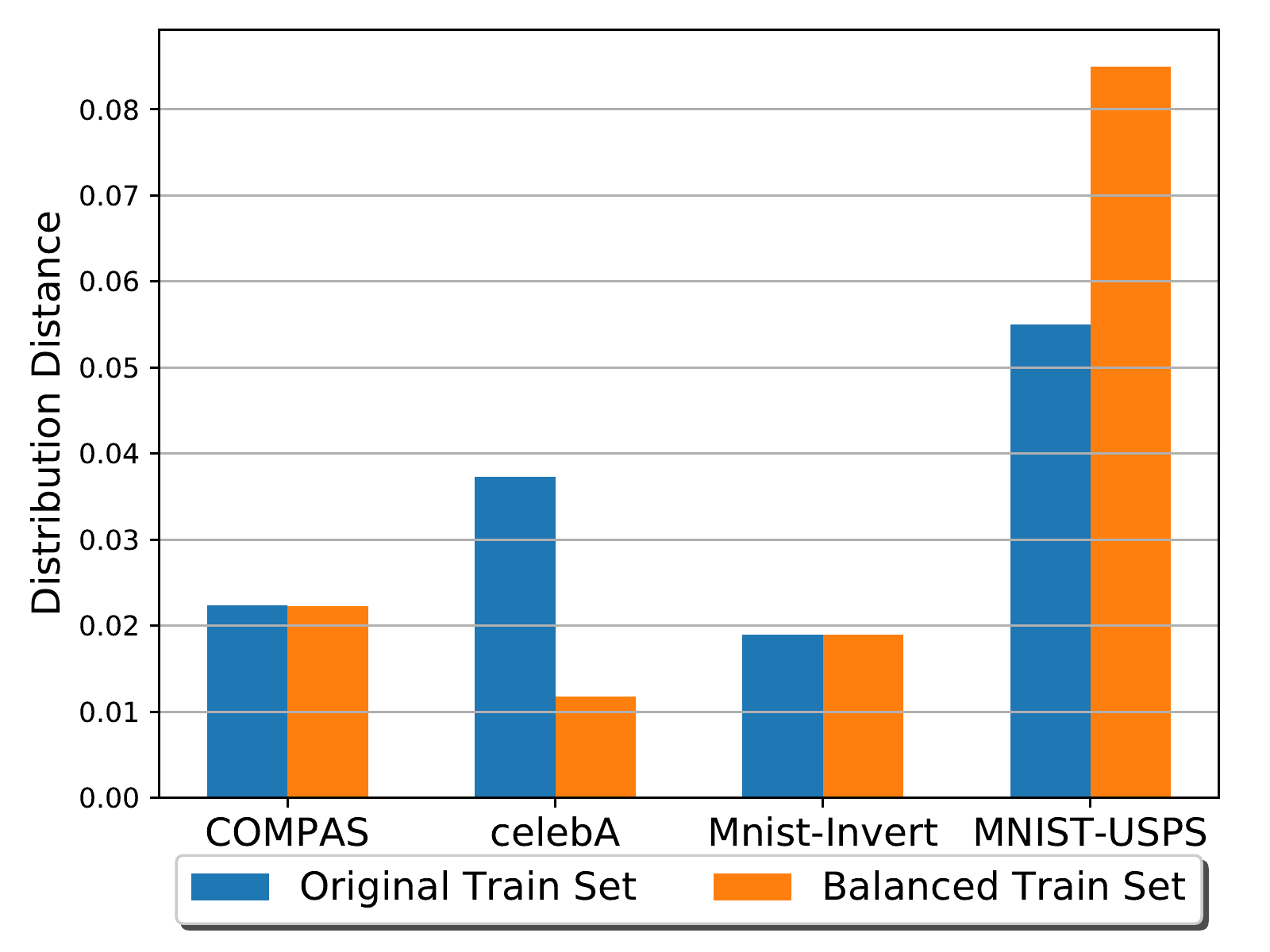}}
 \label{3c}
  \hfill
 \subfloat[DCAE (distribution distance)]{
 \includegraphics[width=0.24\textwidth]{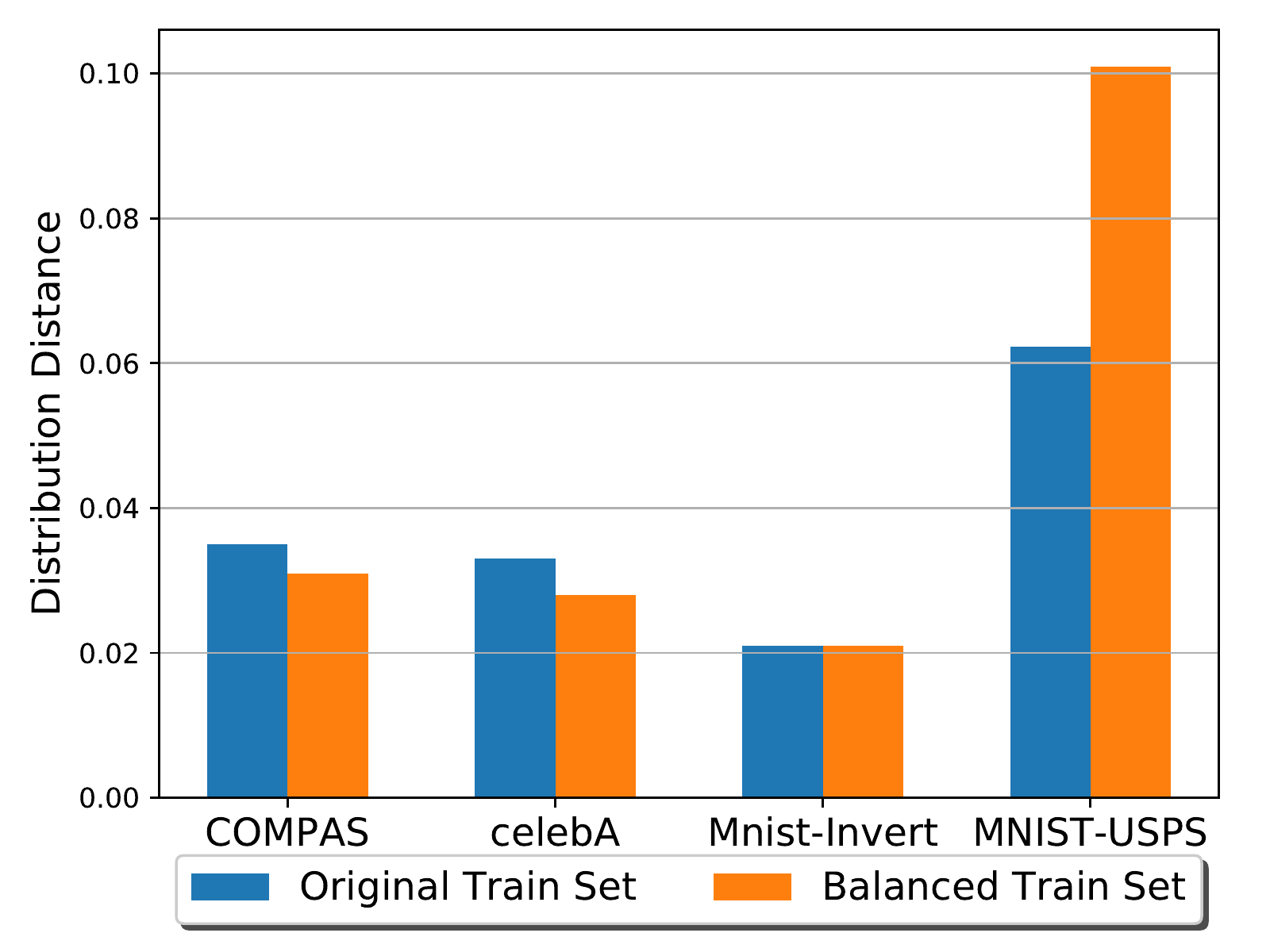}}
 \label{3d}
\caption{Two methods of evaluating the unfairness for existing deep anomaly detection methods on both the original training sets (blue bars) and balanced training sets (orange bars). Note the larger fairness by $p \%$ -rule and smaller distribution distances means the model is fairer. Observed from these figures we can see that training deep anomaly detection models with a balanced training set can slightly improves the fairness in most cases. However, in most cases the fairness by $p \%$ -rule do not satisfy the $80\%$ rule (black horizontal line) advocated by the US Equal Employment Opportunity Commission \cite{biddle2006adverse}.}
\label{fig:svdd_fariness}
\end{figure*}
\subsection{Evaluation Metrics and Baselines}
\label{sec:evaluation_metric}
In our experiments, we evaluate two aspects of the proposed approaches and the baseline methods. The first aspect is the ability to detect anomalies. We evaluate the anomaly detection performance using the common Area Under the ROC Curve (AUC). The AUC measure can be thought of as the probability that an anomalous example is given a higher anomaly score than a normal example. In this way, the higher the AUC score is better. The benefit of using AUC is because it represents the anomaly detection performance across various anomaly score thresholds $t$. The second aspect is the ability to be fair in terms of protected status variables. We use aforementioned $p \%$ -rule (equation \ref{eq:fairness}) and distribution distance (equation \ref{eq:distribution_distance}) measures as our evaluation metrics. 

We compare deep fair SVDD with two popular deep anomaly detection methods: deep SVDD \cite{ruff2018deep} and deep convolutional auto-encoders (DCAE) \cite{masci2011stacked}. We duplicate the deep fair SVDD's encoder network architecture for those two deep anomaly detection baselines to make a fair comparison. We use the default parameters suggested in their original papers.

 \begin{figure*}[th]
  \vskip -0.03in
 \centering
 \subfloat[Fairness by $p \%$ -rule ]{
 \includegraphics[width=0.32 \textwidth]{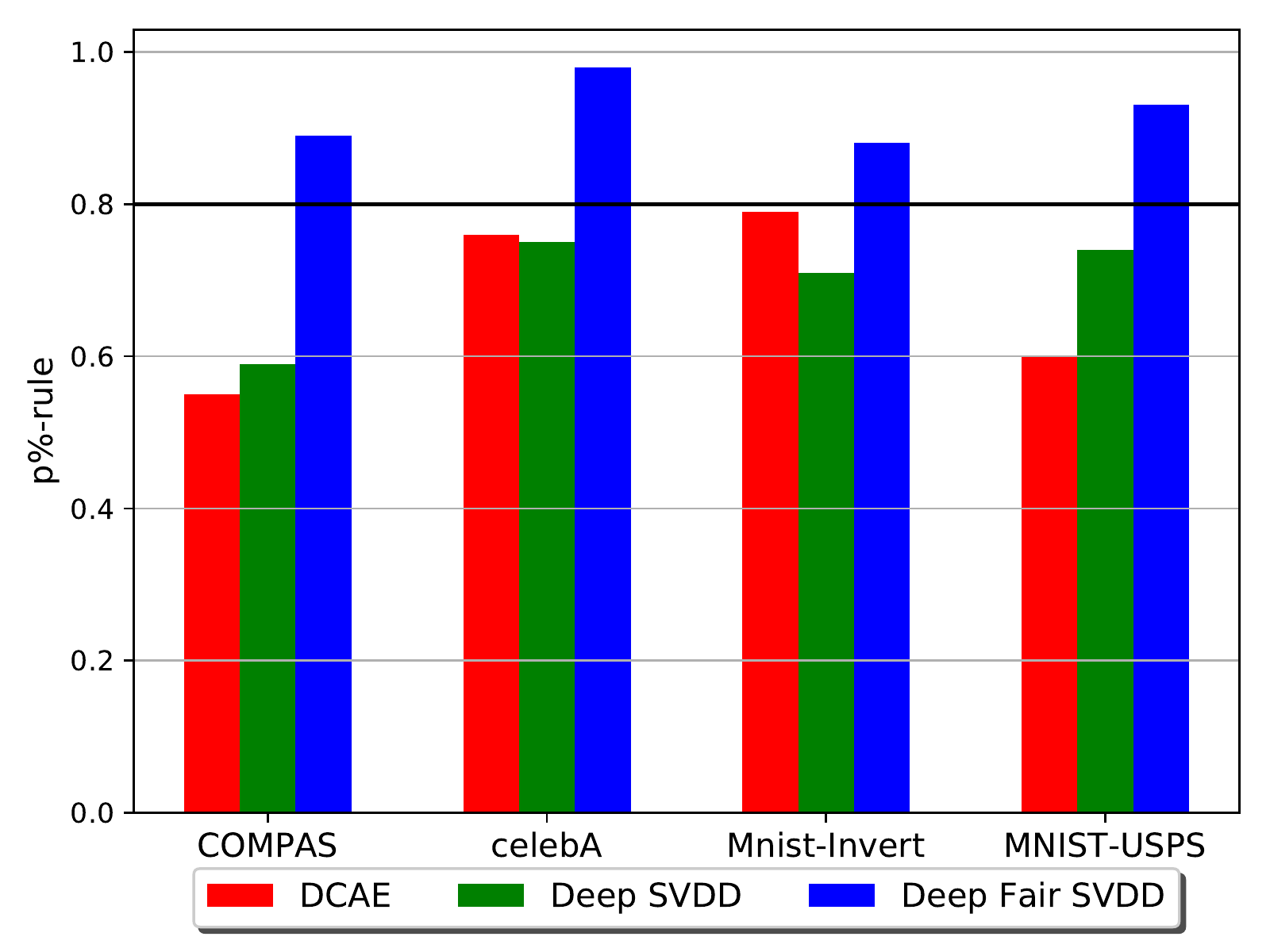}}
 \label{4a}
 \hfill
 \subfloat[Fairness by distribution distance ]{
 \includegraphics[width=0.32\textwidth]{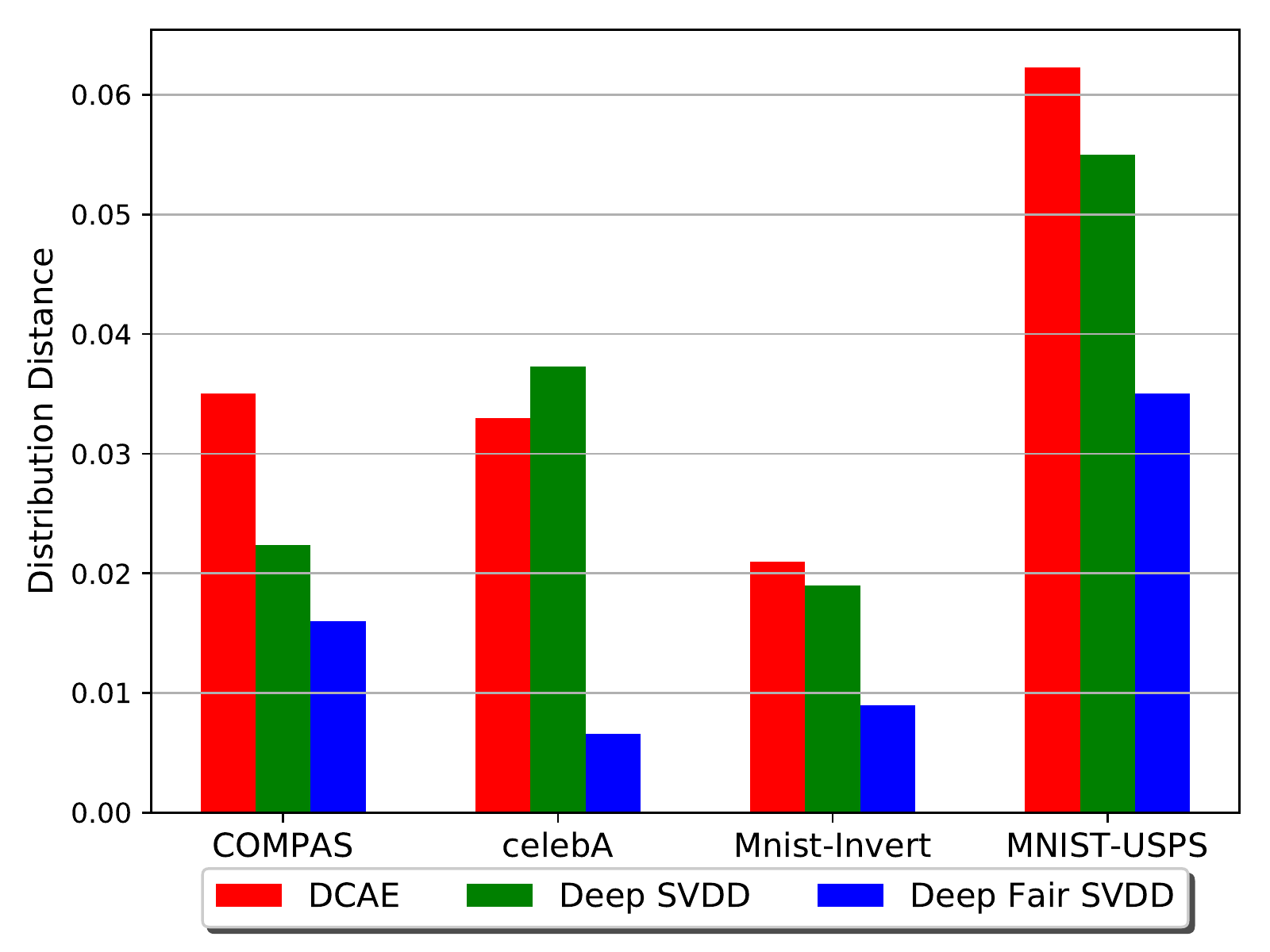}}
 \label{4b}
  \hfill
 \subfloat[AUC Scores]{
 \includegraphics[width=0.32\textwidth]{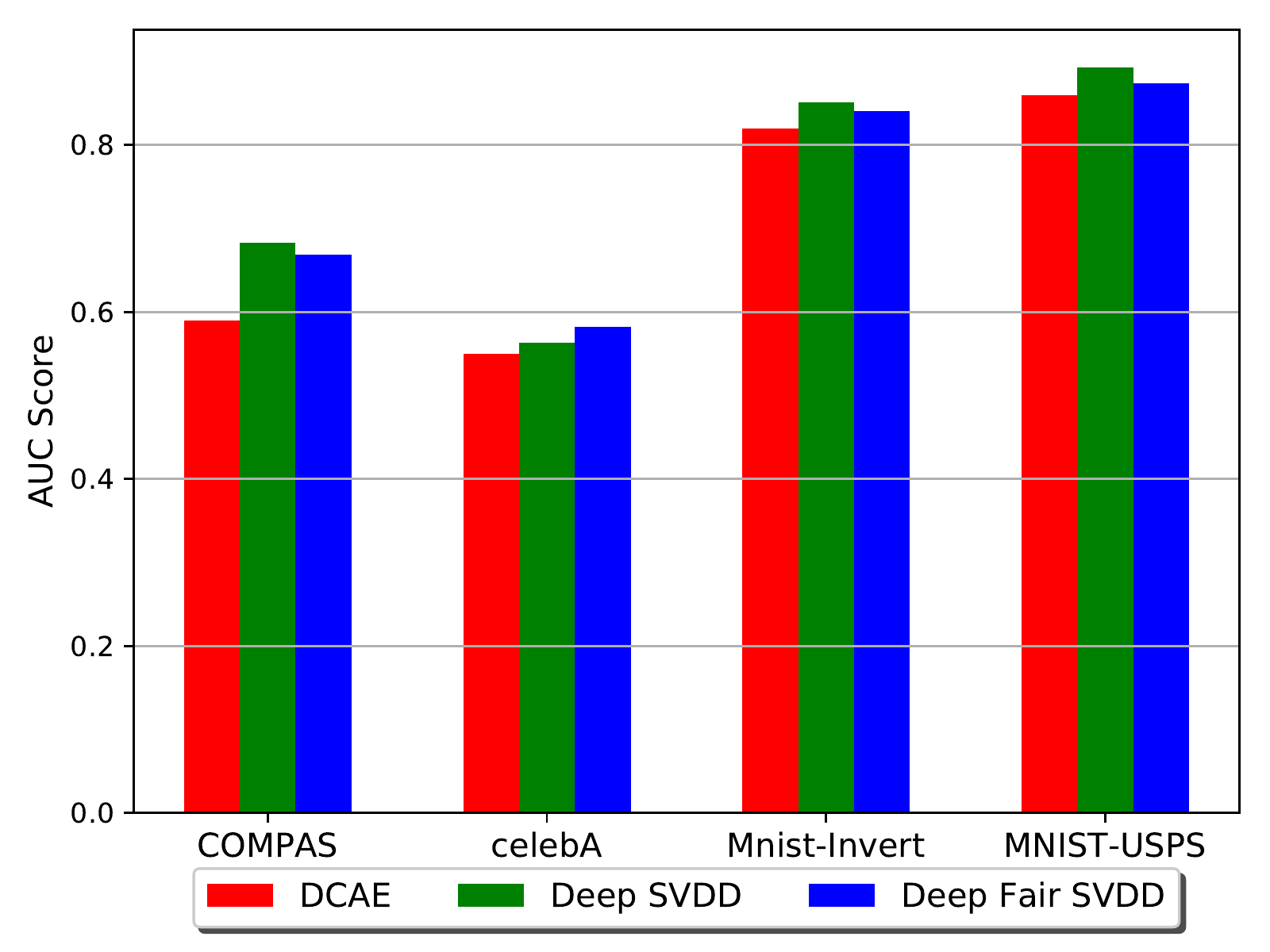}}
 \label{4c}
\caption{Comparison of deep fair SVDD with deep anomaly detection baseline methods on all four selected data sets. We evaluate the fairness performance for all the models trained on original data sets and plot the fairness by $p \%$ -rule and distribution distances in Figure (a), (b). We also evaluate the anomaly detection performance and show the AUC scores in Figure (c). Note deep fair SVDD achieves better fairness results with a slightly loss in terms of the AUC score.} 
\label{fig:fair_svdd_fariness}
\end{figure*}

 \begin{figure*}[h]
 \centering
 \subfloat[Deep SVDD (MNIST-Invert)]{
 \includegraphics[width=0.24\textwidth]{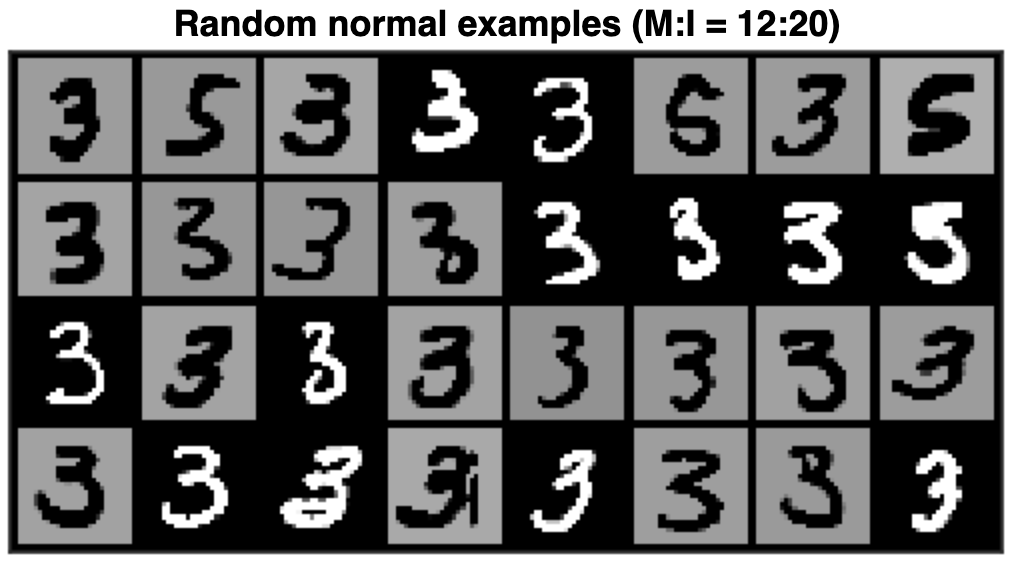}}
 \label{5a}
 \hfill
 \subfloat[Deep SVDD (MNIST-Invert)]{
 \includegraphics[width=0.24\textwidth]{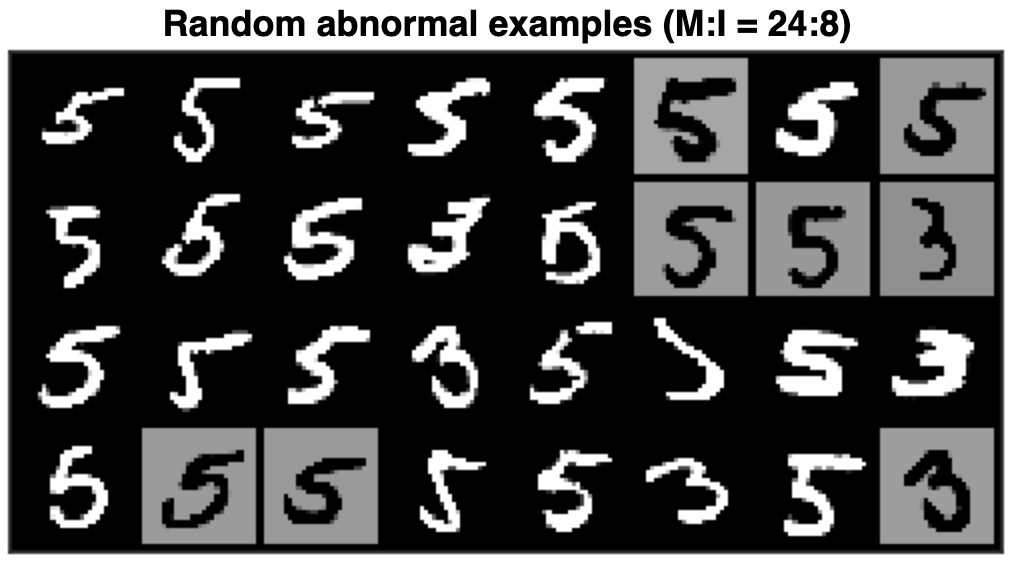}}
 \label{5b}
  \hfill
 \subfloat[Deep SVDD (celebA)]{
 \includegraphics[width=0.24\textwidth]{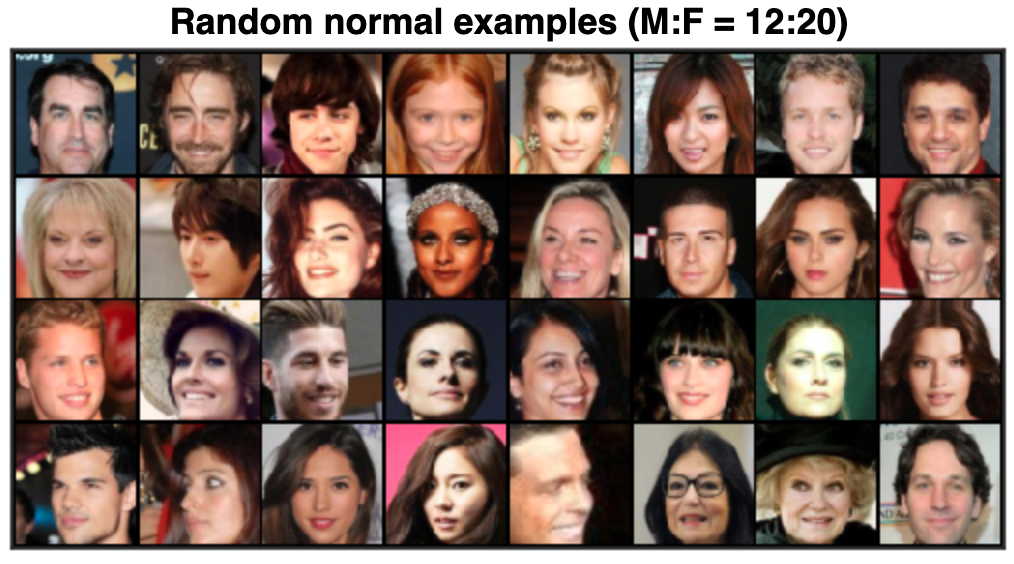}}
 \label{5c}
  \hfill
 \subfloat[Deep SVDD (celebA)]{
 \includegraphics[width=0.24\textwidth]{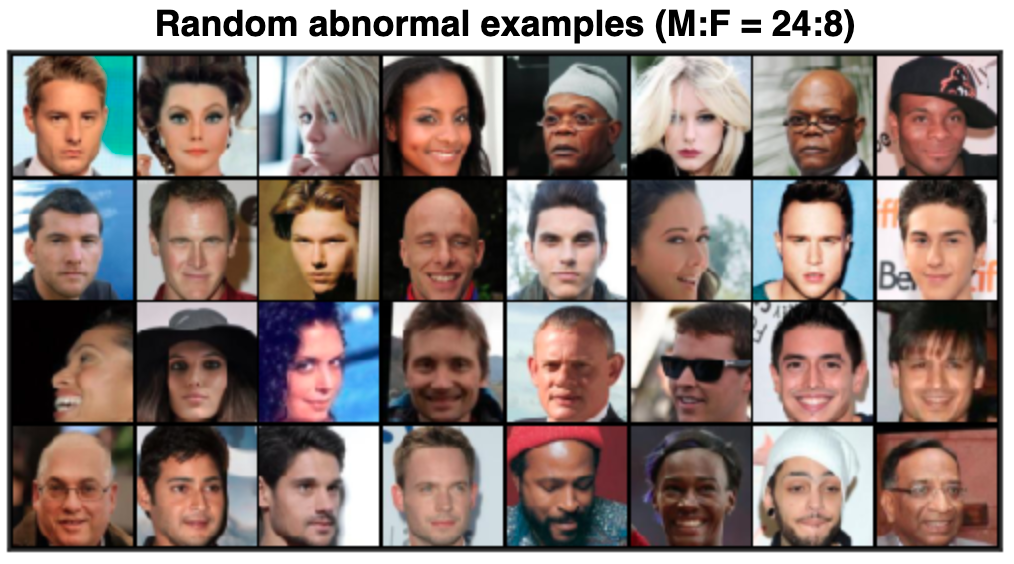}}
 \label{5d}
 \hfill
 \subfloat[Deep Fair SVDD (MNIST-Invert)]{
 \includegraphics[width=0.24\textwidth]{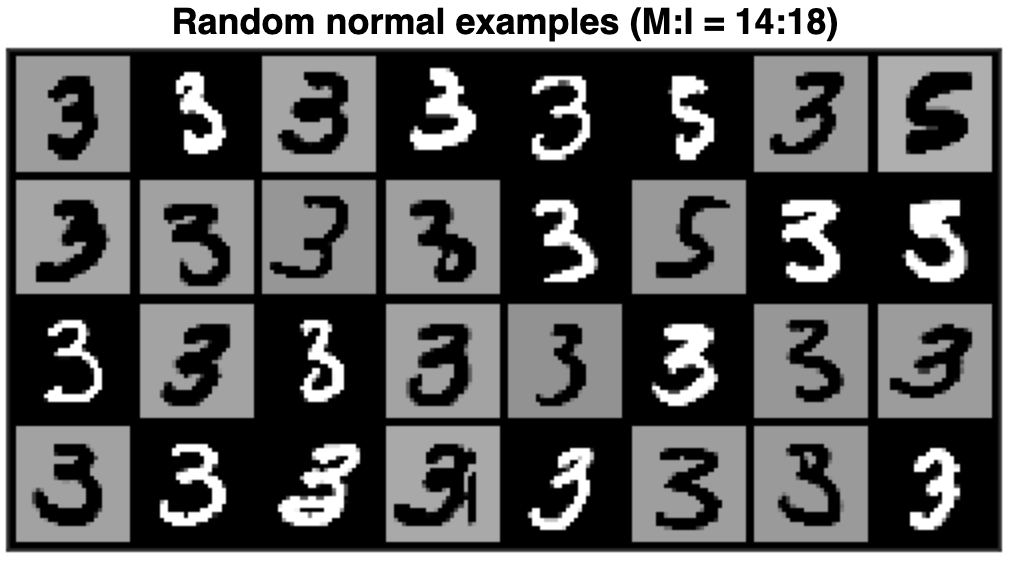}}
 \label{5e}
 \hfill
 \subfloat[Deep Fair SVDD (MNIST-Invert)]{
 \includegraphics[width=0.24\textwidth]{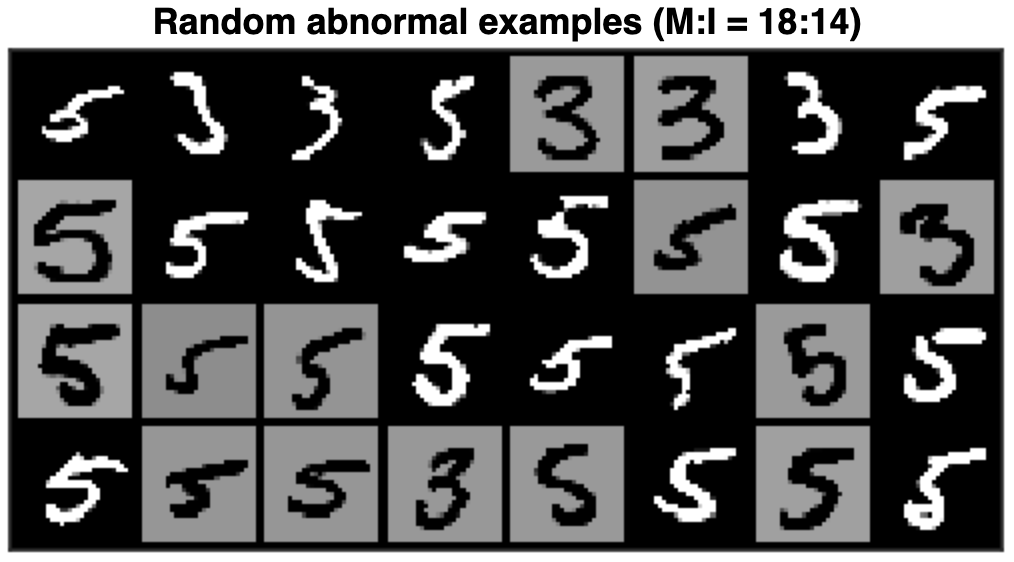}}
 \label{5f}
  \hfill
 \subfloat[Deep Fair SVDD (celebA)]{
 \includegraphics[width=0.24\textwidth]{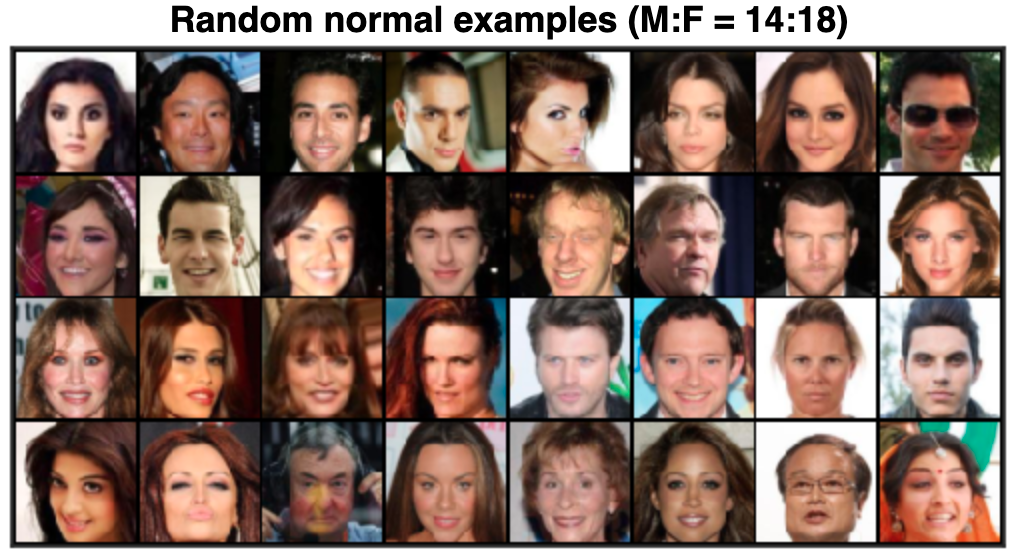}}
 \label{5g}
  \hfill
 \subfloat[Deep Fair SVDD (celebA)]{
 \includegraphics[width=0.24\textwidth]{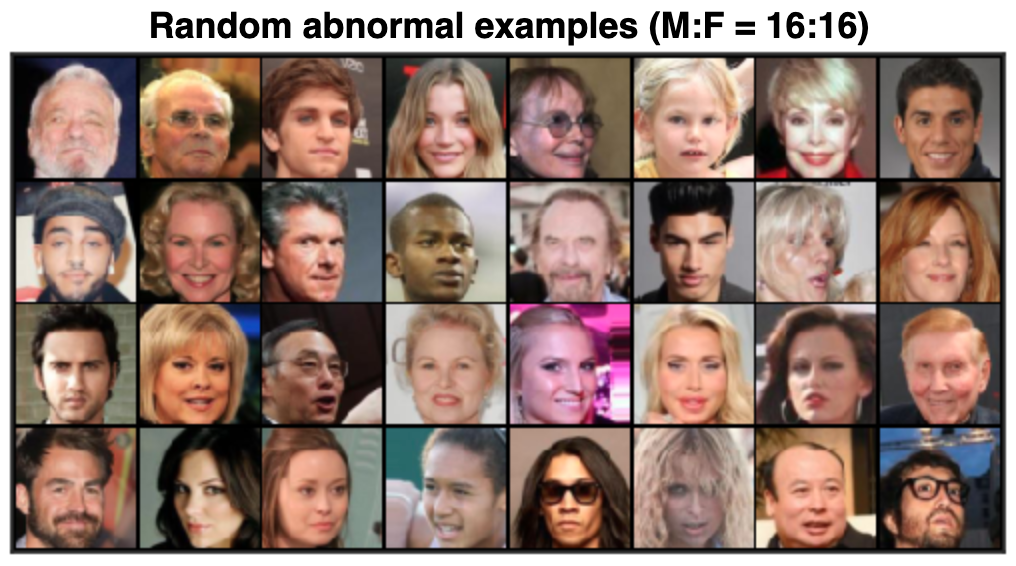}}
 \label{5h}
\caption{The visualization of the random selected normal and abnormal examples determined by deep SVDD (top row) and deep fair SVDD (bottom row) for MNIST-Invert data set and celebA data set. Comparing to the deep SVDD's prediction results, the size of instances with different protected status variable values are more balanced in fair SVDD's predictions.}
\label{fig:fair_svdd_example}
\end{figure*}
\subsection{The Unfairness of Deep Anomaly Detection}
\label{sec:unfair_dod}
We first study the problem of whether existing deep anomaly detection methods can generate fair predictions. We study this under two settings one where we balance the PSV one where we do not.  An imbalanced data set can very easily lead to unfair results whilst a balanced data set is easier to find fair anomalies. To demonstrate that deep anomaly detection models are unfair, we have prepared two versions of the training set: the original training set and the balanced training set. We have listed the detailed information in table \ref{tab:training_set}. If the deep anomaly detection models can't generate fair predictions with both original and balanced training set, then we can conclude that our selected deep anomaly detection methods are unfair.

Thus, we conduct anomaly detection experiments and report both deep SVDD and DCAE's fairness performance on both versions of training sets in figure \ref{fig:svdd_fariness}. We select these two methods because they represent the two popular types of deep anomaly detection methods. Observing Figure \ref{fig:svdd_fariness} (a) and (b), We can see for both COMPAS and celebA data set the deep SVDD and DCAE achieves higher fairness by $p \%$ -rule with a balanced training set. However, the improvements are not ideal because both approaches only satisfied the $80\%$ rule on one data set (celebA). Moreover, for the MNIST-USPS data set, both deep SVDD and DCAE become more unfair with a balanced training set.  

Figure \ref{fig:svdd_fariness} (c) and (d) shows the distribution distance which reflects the overall fairness of each model. The smaller distances indicate the model's predictions are more likely to be independent with the sensitive attribute. We can observe a similar trend as we have seen in Figure \ref{fig:fair_svdd_fariness} (a) and (b) that learning on a balanced training set can only provide marginal improvements. We learn from these results that a fair anomaly detection approach is needed to mitigate deep anomaly detection algorithms' unfairness.

\subsection{Evaluating Deep Fair SVDD}
\label{sec:fair_svdd}
We now evaluate our proposed deep fair SVDD networks' performance and make a comparison with deep SVDD and DCAE. Figure \ref{fig:fair_svdd_fariness} (a) shows the fairness by $p \%$ -rule on abnormal groups. We can see that deep fair SVDD outperforms both deep SVDD and DCAE in all four data sets. Moreover, deep fair SVDD's fairness by $p \%$ -rule are greater than $80\%$ which satisfies the  $80\%$ rule \cite{biddle2006adverse} advocated by the US Equal Employment Opportunity Commission. The distribution distance results are shown in Figure \ref{fig:fair_svdd_fariness} (b).  We can see that deep fair SVDD achieves better overall fairness performance, especially for the celebA data set. Lastly, we show the test set AUC scores for four data sets in Figure \ref{fig:fair_svdd_fariness} (c); we notice that in COMPAS, MNIST-Invert, and MNIST-USPS data sets, the deep SVDD performs slightly better than the other two approaches, while in the celebA data set the deep fair SVDD performs slightly better than other two approaches. Overall speaking, deep fair SVDD achieves much better fairness with a minimal loss in anomaly detection performance. Further, we analyze the interesting result on the celebA data set. In the celebA test set, both the normal and abnormal groups have a balanced number of males and females. Thus optimizing fairness in the celebA data set may also improve the anomaly detection performance. We have observed similar results in the following experiments on the trade-off analysis of deep fair SVDD (\ref{sec:trade_off}).

 \begin{figure*}[th]
 \centering
 \subfloat[COMPAS]{
 \includegraphics[width=0.24\textwidth]{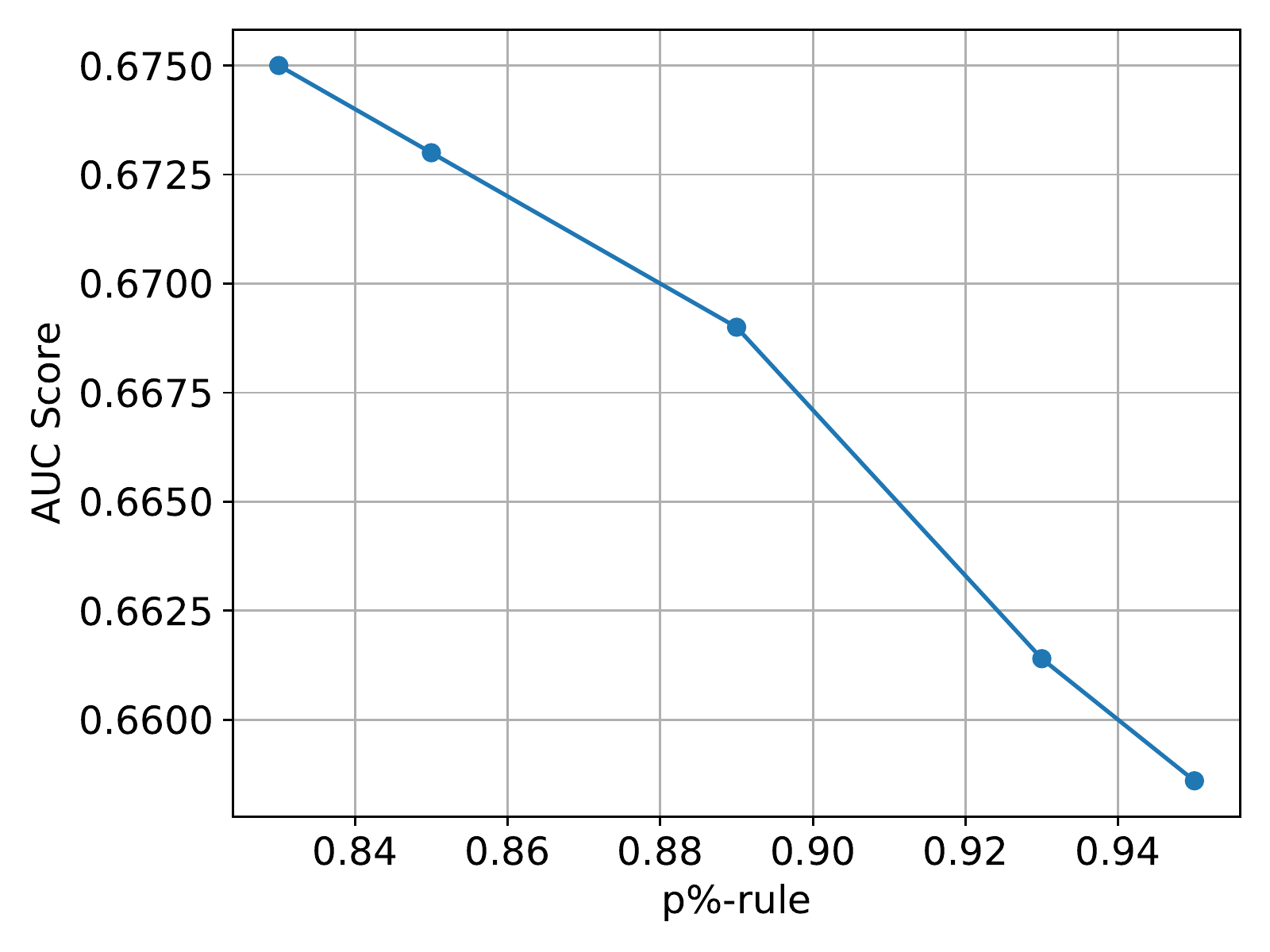}}
 \label{6a}
 \hfill
 \subfloat[celebA]{
 \includegraphics[width=0.24\textwidth]{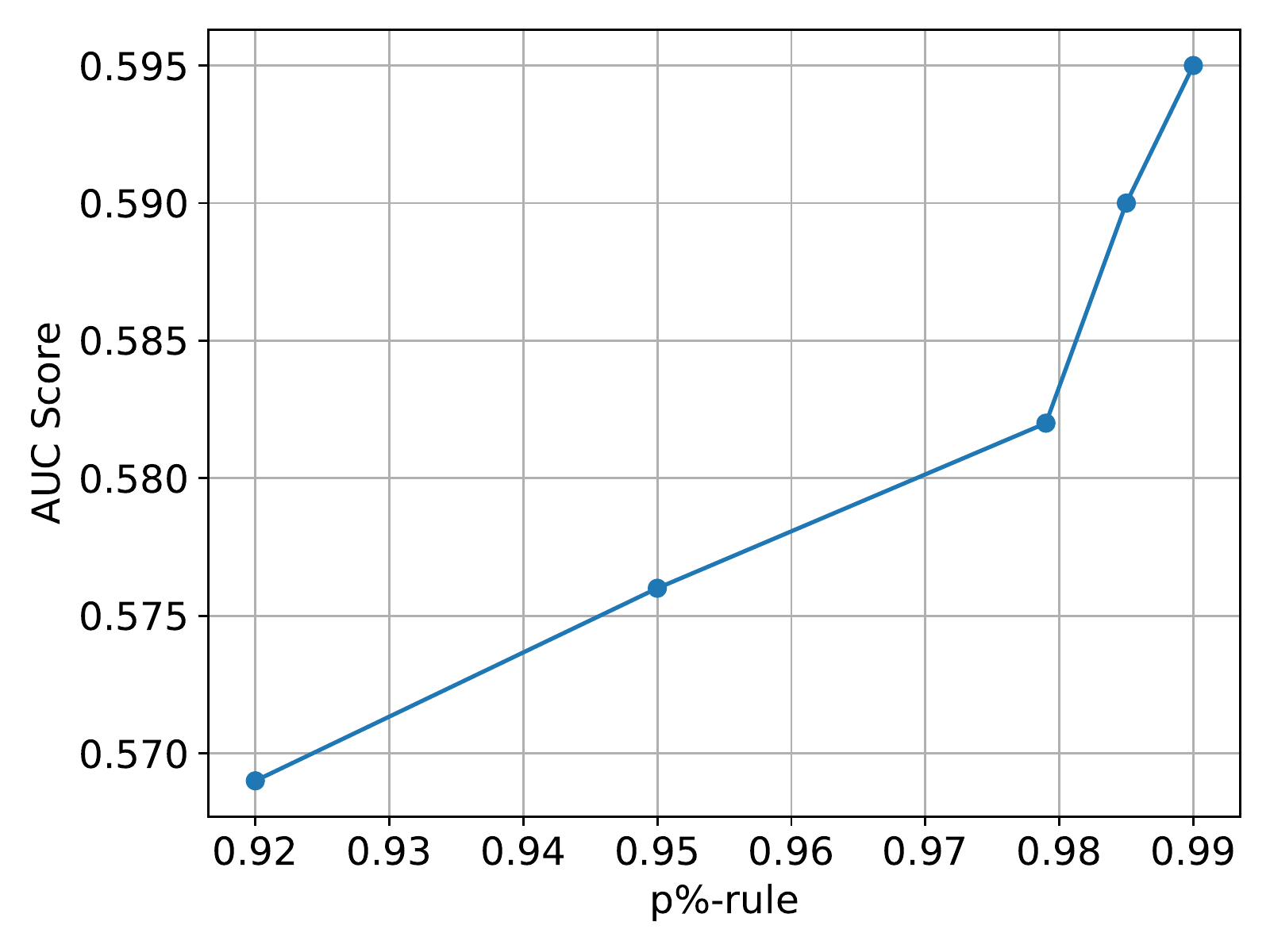}}
 \label{6b}
  \hfill
 \subfloat[MNIST-Invert]{
 \includegraphics[width=0.24\textwidth]{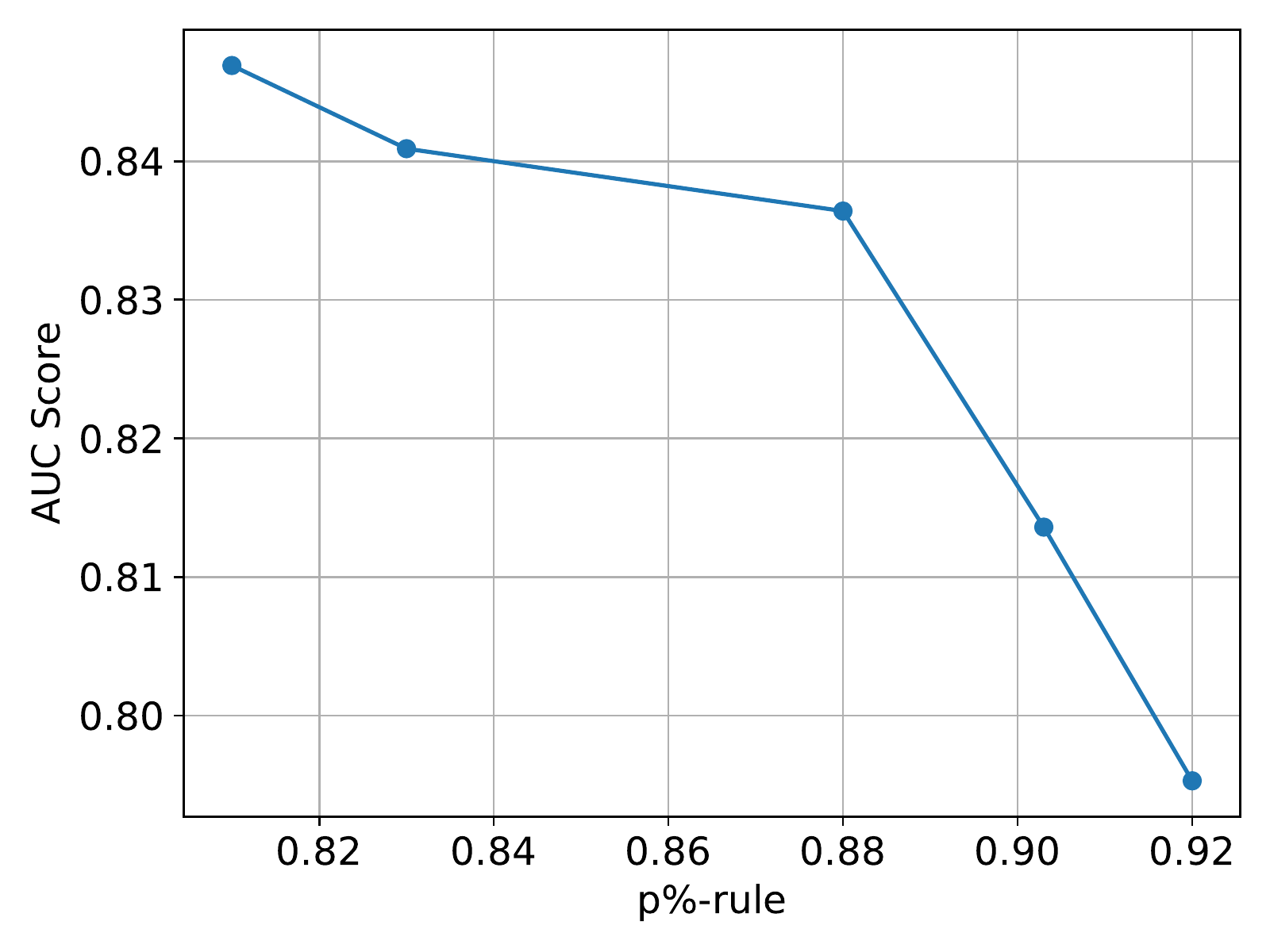}}
 \label{6c}
  \hfill
 \subfloat[MNIST-USPS]{
 \includegraphics[width=0.24\textwidth]{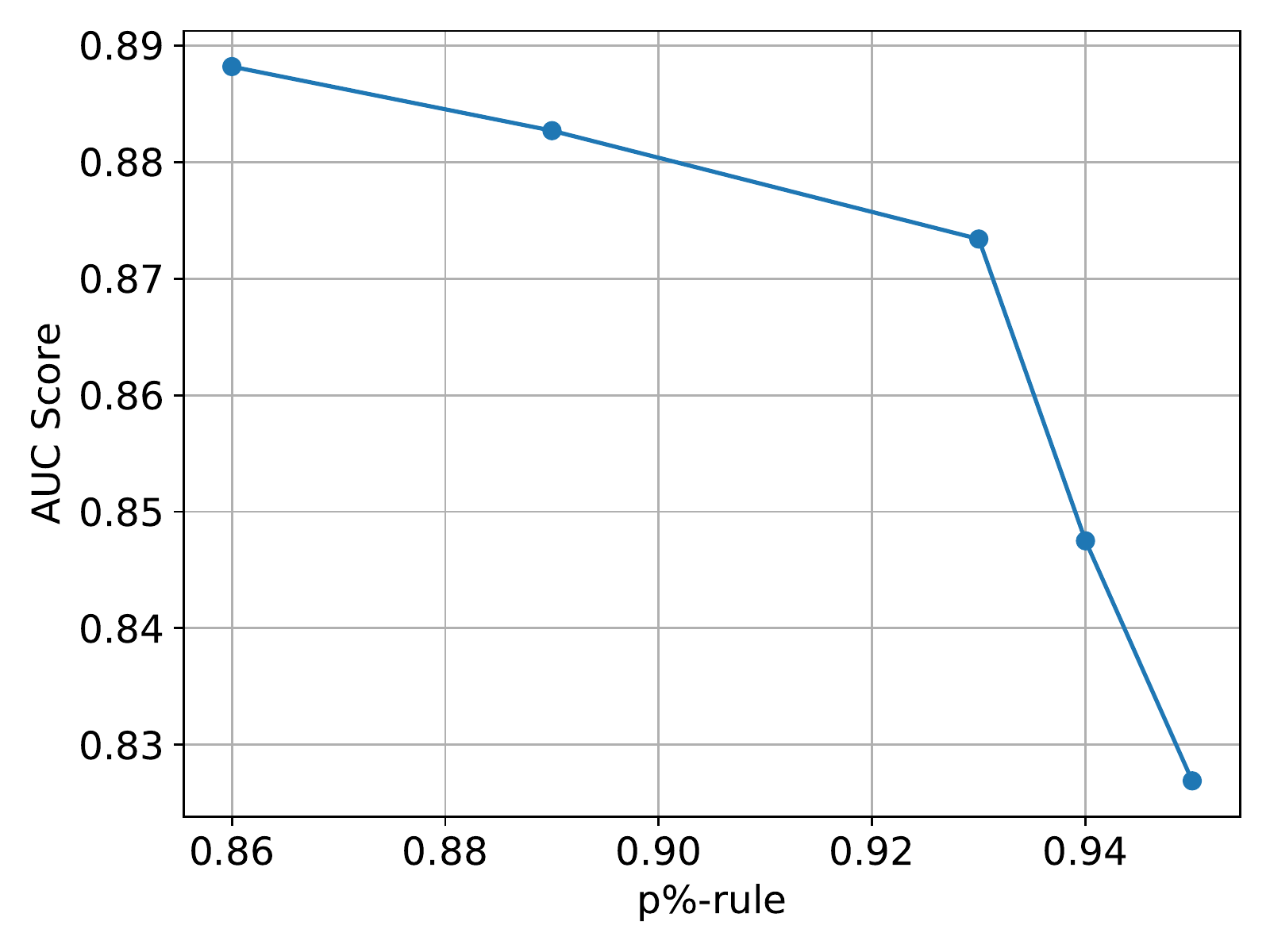}}
 \label{6d}
 \caption{The trade-off between fairness and anomaly detection performance. We tune the hyper-parameter $\lambda$ to demonstrate the trade-off between fairness by $p \%$ -rule and anomaly detection performance in all the data sets. Note the $\lambda$ ranges from ${10} ^{-2}$ to $10^{2}$ and it is visualized in each plot with the order from left to right respectively. In all four datasets the fairness by $p \%$ -rule value increases as $\lambda$ increases. The AUC scores decrease in most data sets as $\lambda$ increases.}
\label{fig:trade-off}
 \end{figure*}

Figures \ref{fig:fair_svdd_example} shows examples of the random selected normal and anomalous examples according to deep SVDD and deep fair SVDD's predictions. For the MNIST-Invert data set, we can see that both the MNIST instances and Inverted MNIST instances are distributed evenly in the normal/abnormal groups determined by deep fair SVDD. On the contrary, there are more MNIST instances in the abnormal group and fewer MNIST instances in the normal group determined by deep SVDD. As for the anomaly detection quality, both approaches have made few mistakes and achieved similar results, as shown in Figure \ref{fig:fair_svdd_fariness}. 

The right-hand side of the Figure \ref{fig:fair_svdd_example} shows the results for the celebA data set. Observing the deep SVDD's results on the top row shows that more males are predicted as plain faces and more females are predicted as attractive faces. These unfair results are mitigated with deep fair SVDD and we can see a nearly balanced number of males and females in both groups predicted via fair SVDD. As for the anomaly detection quality, both approaches made some mistakes and these are in line with the AUC scores we have reported in Figure \ref{fig:fair_svdd_fariness} (c). This is reasonable as human faces contain far more information than digits. The anomaly detection tasks over human faces are more challenging than recognizing digits. Our main goal is to demonstrate how deep fair SVDD mitigates the unfair problems caused by deep anomaly detection baselines.

\subsection{The Trade-off between Fairness and Anomaly Detection Performance}
\label{sec:trade_off}
This section analyzes the trade-off between fairness performance and anomaly detection performance of deep fair SVDD. We re-train and test the deep fair SVDD under different values of hyper-parameter $\lambda$ (range from ${10} ^{-2}$ to $10^{2}$) within equation (\ref{adv_loss}). The hyper-parameter $\lambda$ controls the weight of the discriminator's loss term within the adversarial loss function and directly determines the trade-off between the fairness performance and anomaly detection performance. Figure \ref{fig:trade-off} shows the results: in all four selected data sets, the fairness by $p \%$ -rule increases as $\lambda$ increases. The AUC score drops as the fairness by $p \%$ -rule value goes up for COMPAS, MNIST-Invert, and MNIST-USPS data sets. We have also noticed one different result in the celebA data set, both fairness by $p \%$ -rule and AUC score increase as the $\lambda$ increases. We have analyzed this case before when comparing deep fair SVDD to deep anomaly detection baselines in plot \ref{fig:fair_svdd_fariness} (c). Here fairness constraint is extra information that could help the algorithm improve anomaly detection performance. Generally speaking, training the deep fair SVDD with a larger $\lambda$ will lead to fairer results and usually a slight loss in terms of the anomaly detection performance (AUC score).

\subsection{Anomaly Predictions Analysis}
\label{sec:outlier_predictions}
This section will conduct experiments to study how deep fair SVDD's predictions differ from deep SVDD's predictions. We have stored the anomaly prediction results for both approaches and summarized their overlapped anomaly predictions in Table \ref{tab:overlap_prediction}. We use the number of overlapped anomaly predictions to divide the total number of anomalies as the overlap ratio. We can see that the overlap ratios are pretty high across all the data sets. We hypothesize the reason is that fair SVDD is also optimized with SVDD loss function. Furthermore, this high overlapping can also explain why fair SVDD only performs slightly worse than SVDD in terms of the AUC scores as we demonstrated in Figure \ref{fig:fair_svdd_fariness} (c).

\begin{table}[t]
  \caption{ Anomaly prediction results for deep SVDD and deep fair SVDD. $Z_0$ and $Z_1$ represent the number of predicted anomalies with protected status variable value as $0$ and $1$ respectively. There is a large overlap between these two model's anomaly predictions.}
  \label{tab:overlap_prediction}
  \small
  \begin{tabular}{c|c|c|c|c}
    \toprule
         &COMPAS & celebA & MNIST-Invert &MNIST-USPS \\
    \midrule
    SVDD ($Z_{0}$ : $Z_{1}$)  & 198:336    & 854:1146  &   743:1041  & 186:137  \\
    \hline
    Ours ($Z_{0}$ : $Z_{1}$)  & 263:271    & 980:1020   &   832:952  & 164:159  \\
    \hline
    Overlap ratio  & 0.78    & 0.70   &   0.81  & 0.82  \\
  \bottomrule
\end{tabular}
\end{table}

 \begin{figure}[th]
 \centering
 \subfloat[Instances "moved" from normal to abnormal group]{
 \includegraphics[width=0.23 \textwidth]{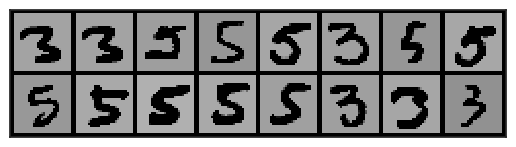}}
 \label{9a}
 \hfill
 \subfloat[Instances "moved" from abnormal to normal group]{
 \includegraphics[width=0.23\textwidth]{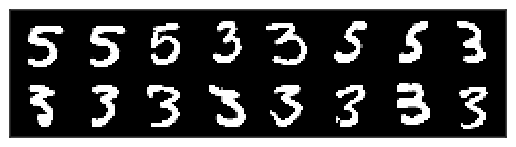}}
 \label{9b}
\caption{Illustration of how deep fair SVDD makes the anomaly detection results fairer. We visualize the sampled non-overlapping predictions between deep SVDD and deep fair SVDD. The instances in (a) can be seen as moved from deep SVDD's predicted normal group to deep fair SVDD's predicted abnormal group and vice versa for (b).}
\label{fig:prediction_visual}
\end{figure}

We also visualize the non-overlapping predictions between deep SVDD and deep fair SVDD in Figure \ref{fig:prediction_visual}. Take the MNIST-Invert data set for example; we randomly sample $16$ non-overlapping anomalies with $z = 0$ from fair SVDD's predictions. We can view these instances as moved from deep SVDD's predicted normal group to deep fair SVDD's predicted abnormal group to make the results fairer. Observing the digits from Figure \ref{fig:prediction_visual} (a), we can see that deep fair SVDD is improving the fairness by moving instances that are "prone to be anomalies" to the abnormal group. One common feature of those instances is that they are dissimilar to a regular style of digit $3$ and many of them are digits $5$. It is important to show that these non-overlapping instances are not randomly distributed but are all prone to be anomalies. This interesting finding demonstrates that our proposed model is optimized to make fair and accurate anomaly predictions instead of random altering predictions to satisfy group-level fairness. We can observe the similar results from Figure \ref{fig:prediction_visual} (b) that instances moved from deep SVDD's abnormal group to deep fair SVDD's normal group are "prone to be normal points."

 \begin{figure*}[th]
 \centering
 \subfloat[COMPAS (Deep SVDD)]{
 \includegraphics[width=0.24\textwidth]{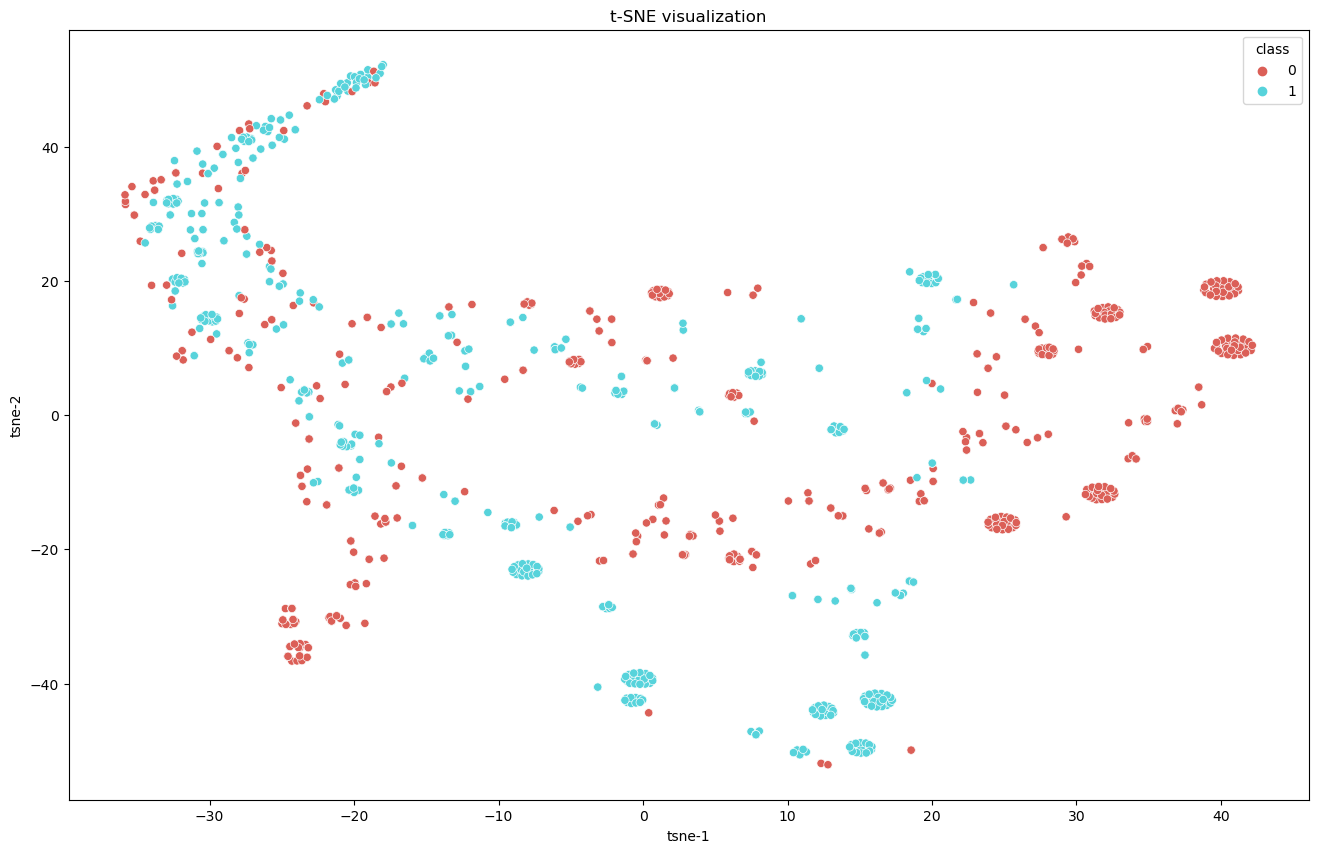}}
 \label{8a}
 \hfill
 \subfloat[celebA (Deep SVDD)]{
 \includegraphics[width=0.24\textwidth]{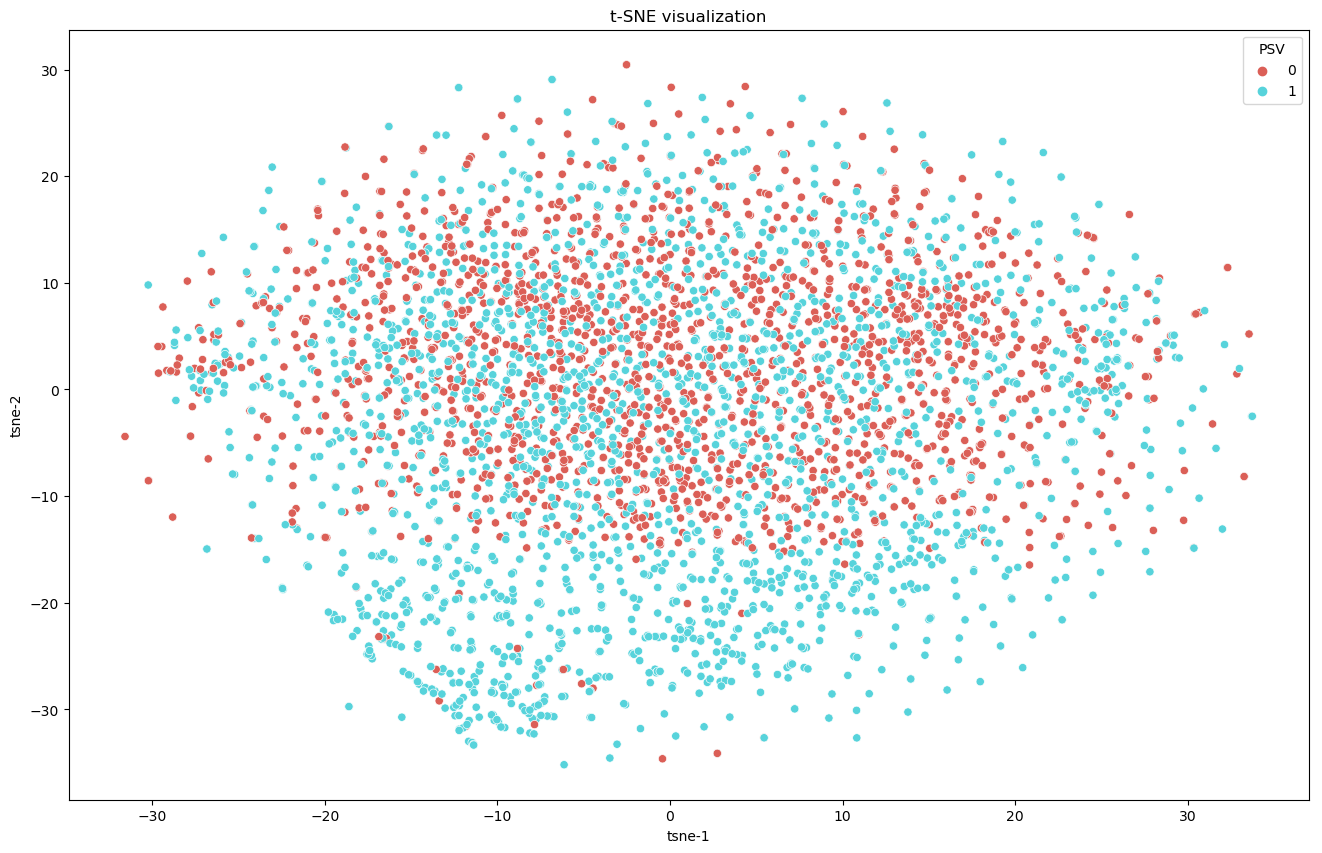}}
 \label{8b}
  \hfill
 \subfloat[MNIST-Invert (Deep SVDD)]{
 \includegraphics[width=0.24\textwidth]{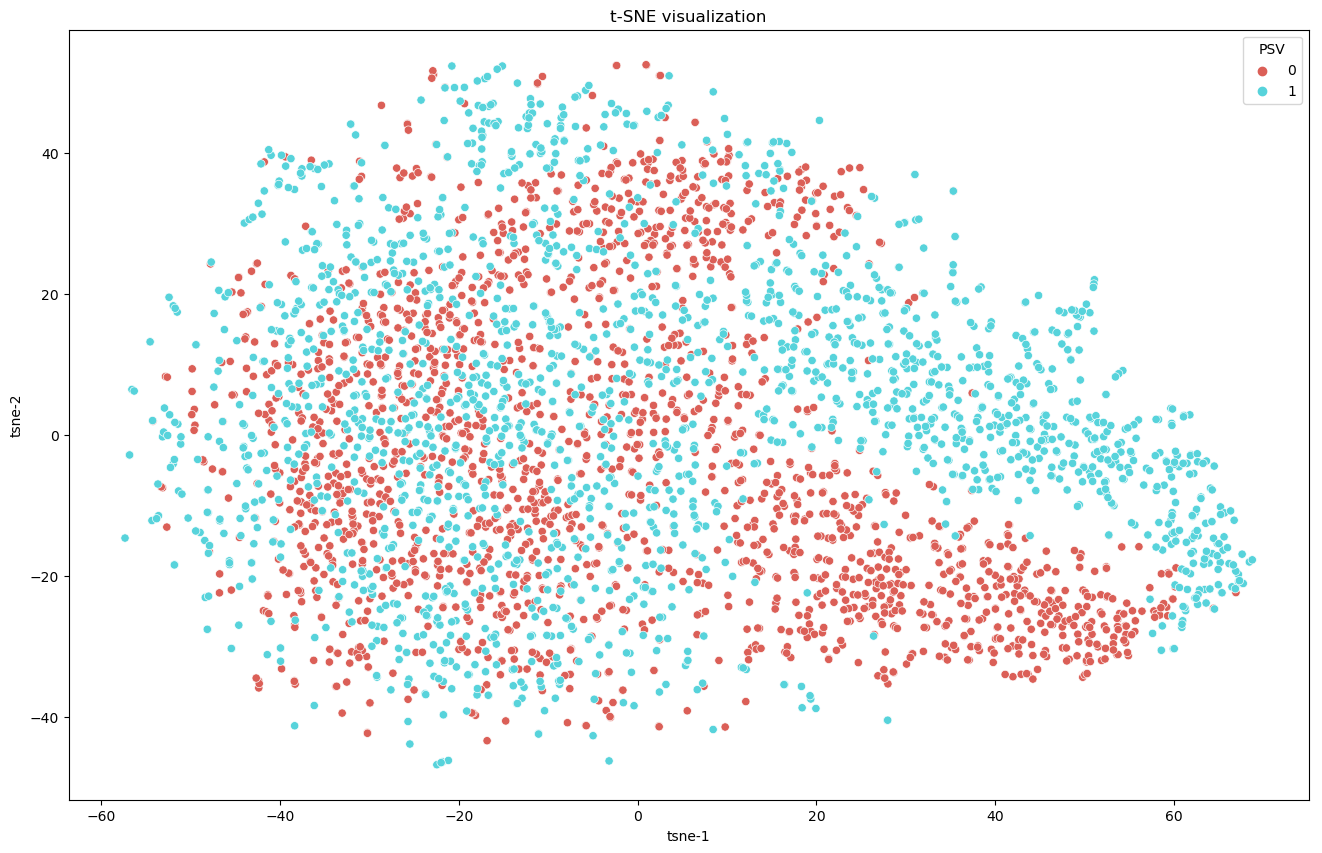}}
 \label{8c}
  \hfill
 \subfloat[MNIST-USPS (Deep SVDD)]{
 \includegraphics[width=0.24\textwidth]{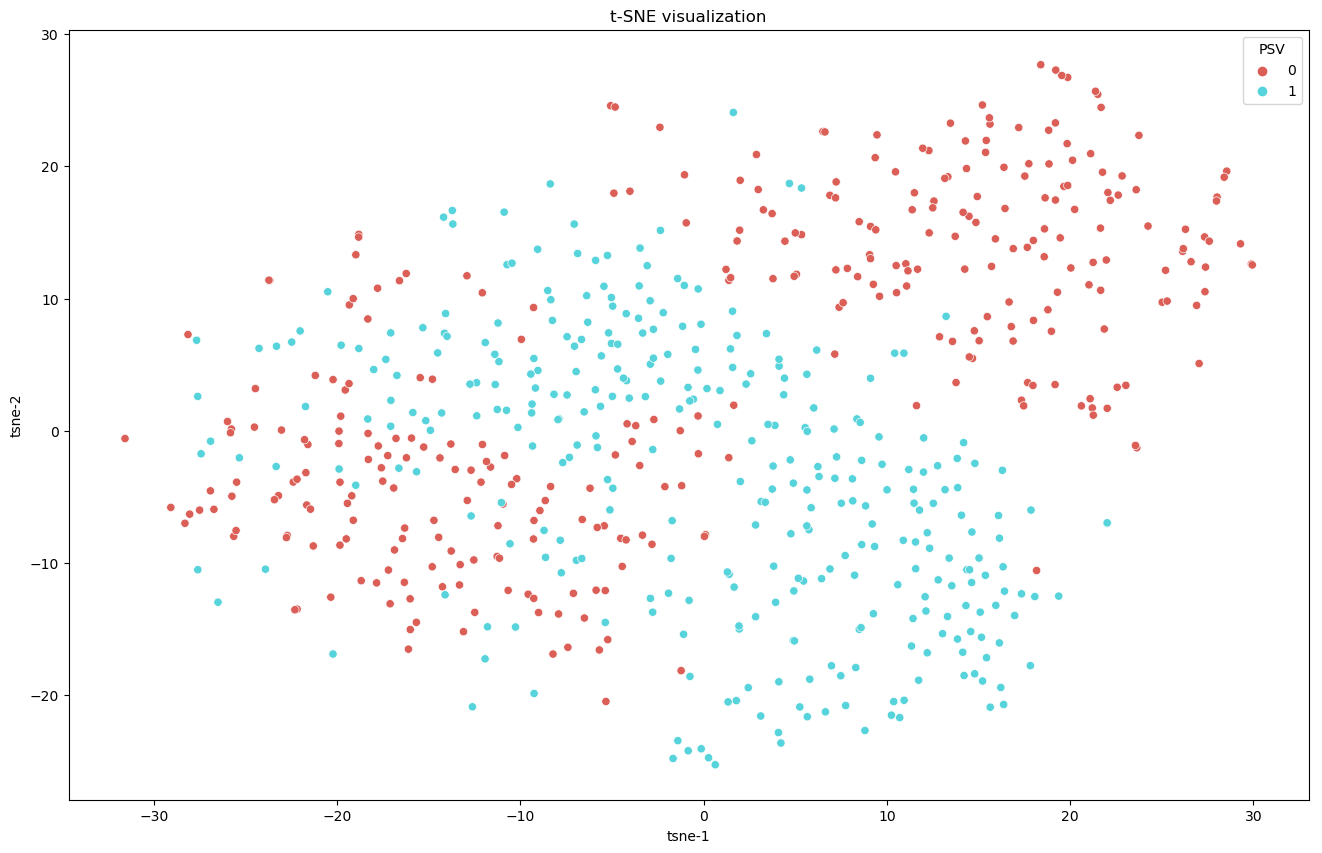}}
 \label{8d}
 \hfill
  \subfloat[COMPAS (Deep Fair SVDD)]{
 \includegraphics[width=0.24\textwidth]{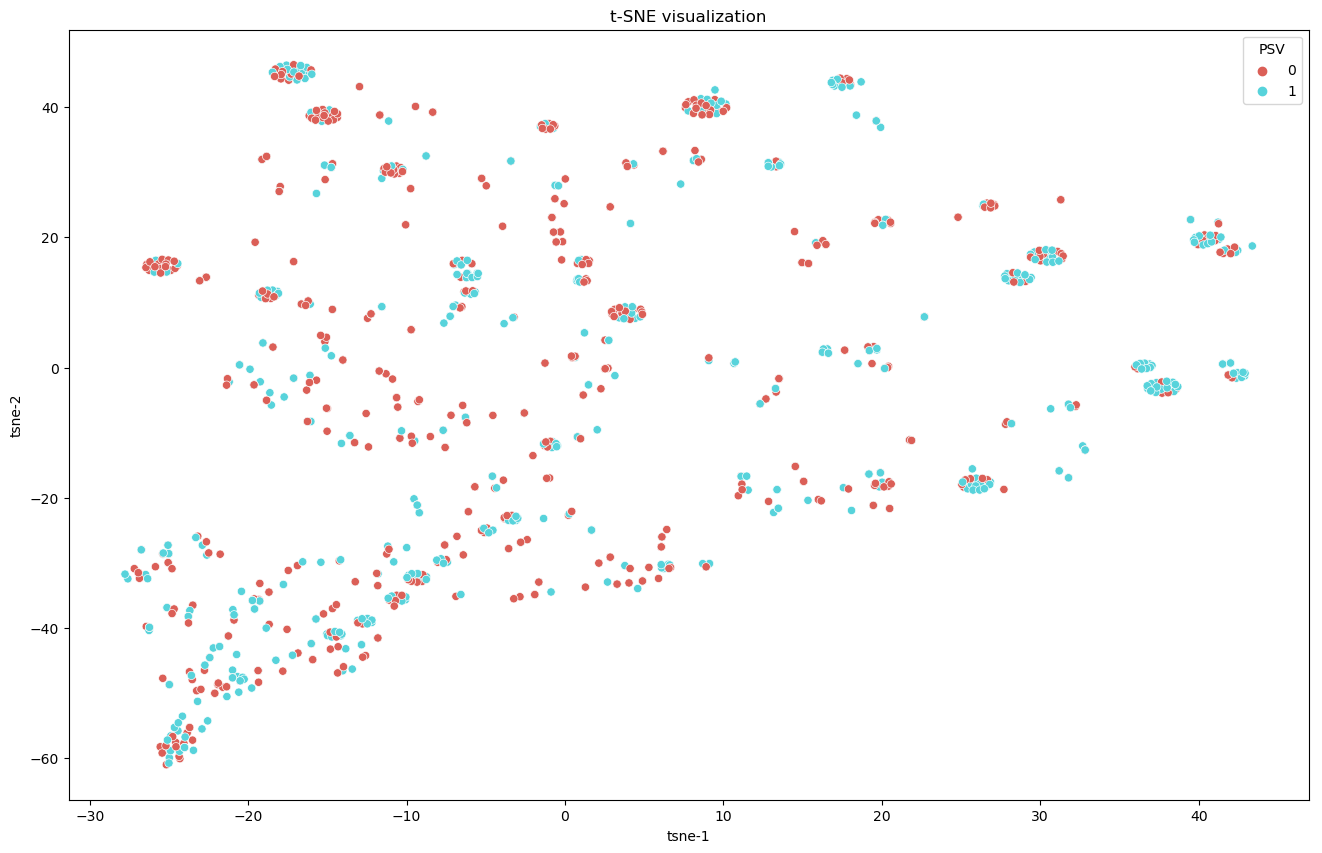}}
 \label{8e}
 \hfill
 \subfloat[celebA (Deep Fair SVDD)]{
 \includegraphics[width=0.24\textwidth]{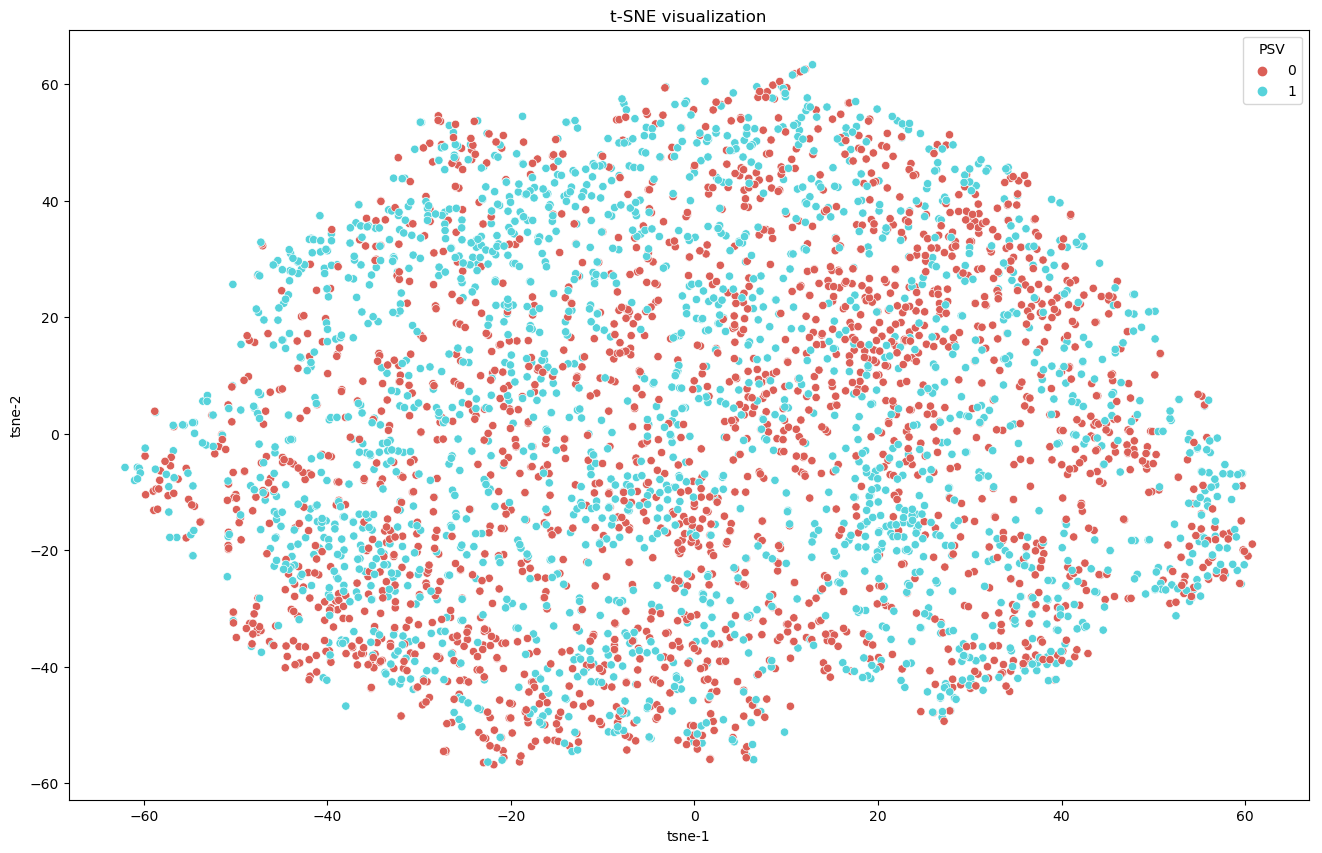}}
 \label{8f}
  \hfill
 \subfloat[MNIST-Invert (Deep Fair SVDD)]{
 \includegraphics[width=0.24\textwidth]{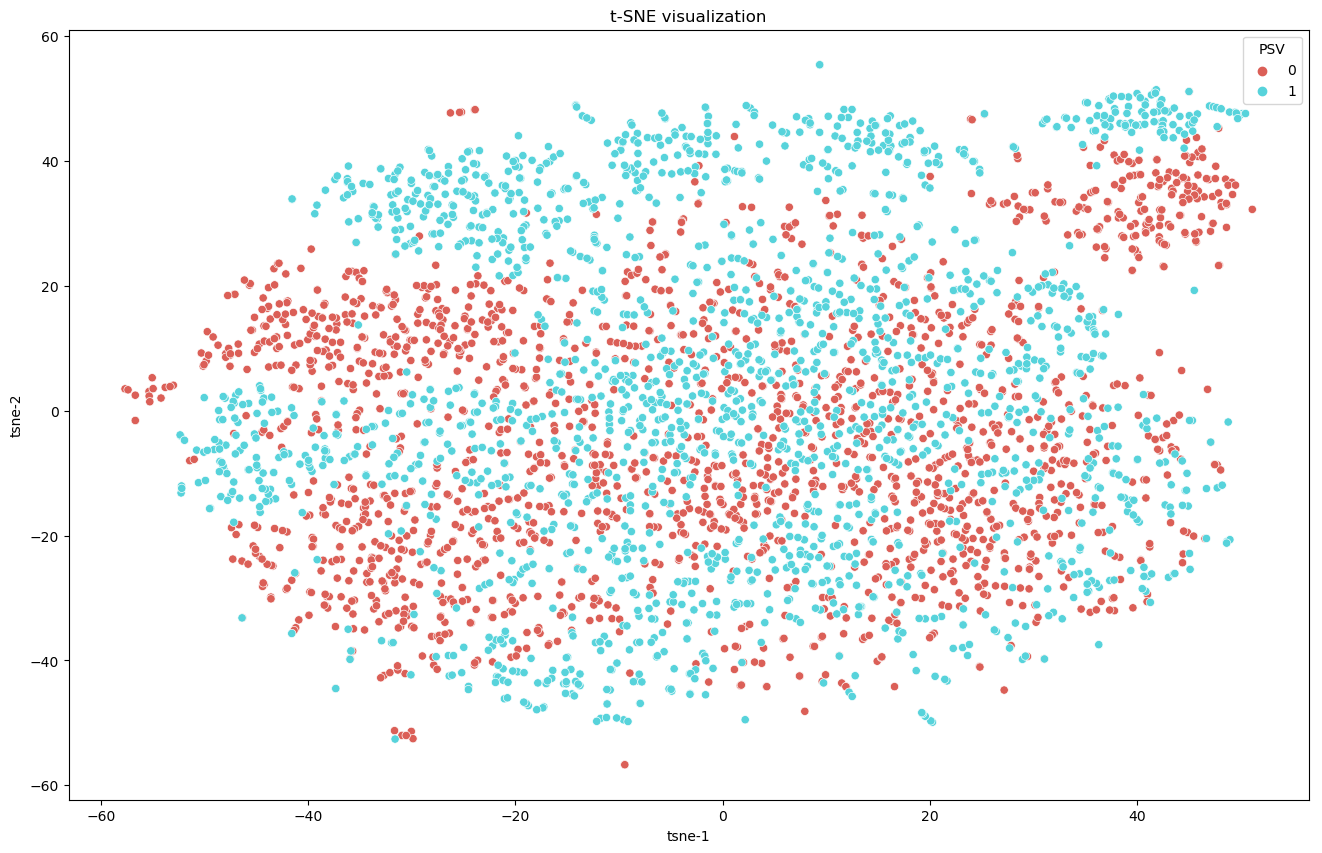}}
 \label{8g}
  \hfill
 \subfloat[MNIST-USPS (Deep Fair SVDD)]{
 \includegraphics[width=0.24\textwidth]{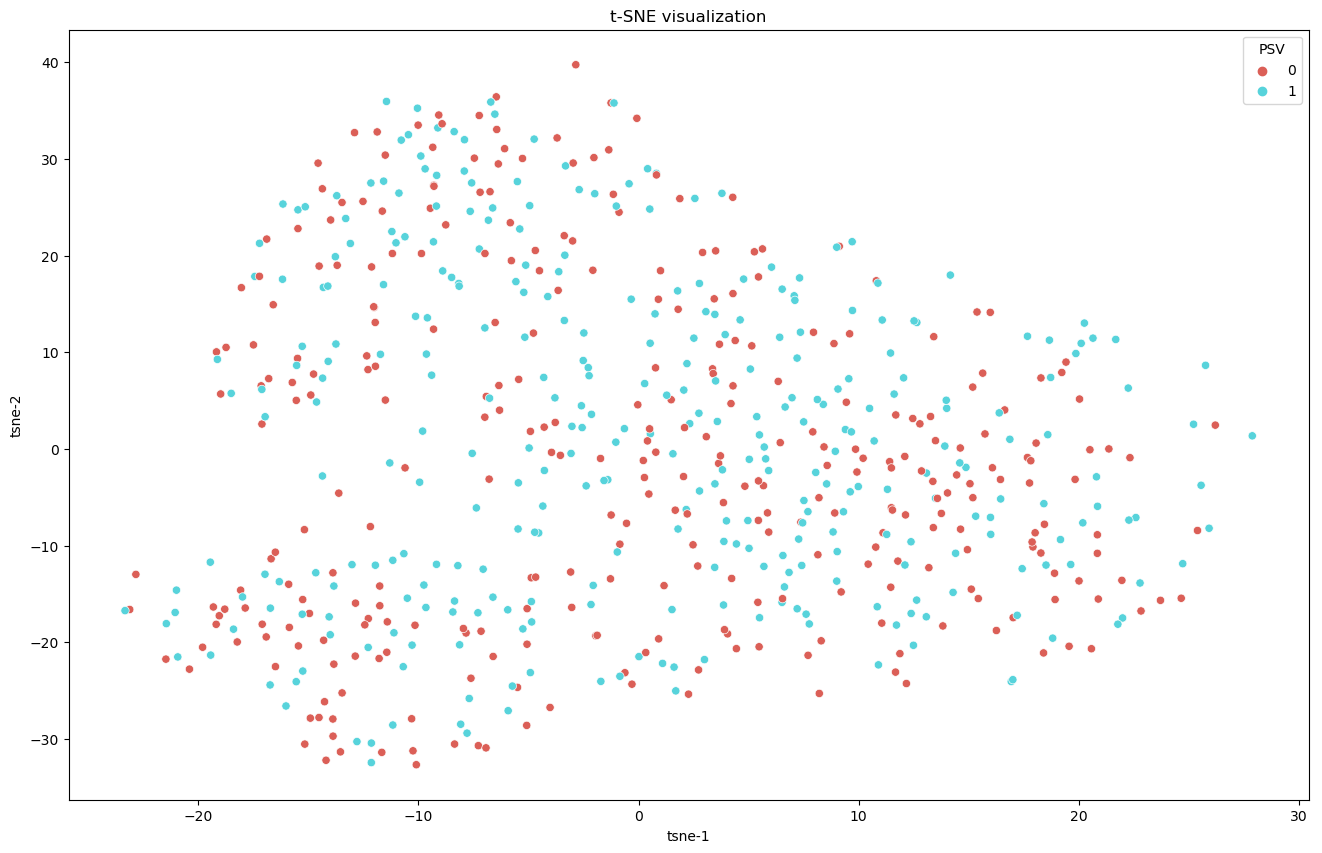}}
 \label{8h}
 \caption{The t-SNE \cite{maaten2008visualizing} visualization of the feature embeddings for test instances. Red and blue points represent test instances with different sensitive attribute values. Comparing to deep SVDD's results (top row), the deep fair SVDD's learned embeddings (bottom row) are more fair as blue and red points are always blended together which are hard to separate. }
\label{fig:embeddings}
 \end{figure*}
 
\subsection{Embedding Visualization}
\label{sec:training_analysis}
We visualize and compare the learned embeddings for both deep SVDD and deep fair SVDD to show why deep fair SVDD make fairer anomaly predictions. This analysis is important as deep fair SVDD's objective is to learn a fair representation which is independent on the protected status variable $z$: $\p (f(\X; \theta) | z = 0 ) = \p (f(\X; \theta) | z = 1)$. 
As shown in Figure \ref{fig:embeddings}, the red and blue points represent the test instances with the sensitive attribute value as $z = 0$ and $z = 1$ respectively. We first analyze the visualization results from deep SVDD; in each plot we can find some regions dominated by one particular color which indicates the correlation between feature embeddings and the protected status variable. On the contrary, observing from the deep fair SVDD's result we can see that the red and blue points are almost uniformly distributed in the feature space especially in the celebA data set. Deep fair SVDD is demonstrated to learn a fair representation that is independent of sensitive attributes.

\subsection{Running Time Analysis}
\label{sec:runtime_analysis}
We have also reported the training time for deep fair SVDD and compared it against the deep SVDD approach in Table \ref{tab:training_time}. Training deep fair SVDD takes longer time because we have a new fairness objective and it is learned through adversarial training. We leave how to speed up the training process as an interesting future work. 
\begin{table}[t]
  \caption{ Training time results measured by seconds. Training deep fair SVDD takes longer time due to the min-max optimization of the adversarial learning.}
  \label{tab:training_time}
  \small
  \vskip -0.1in
  \begin{tabular}{c|c|c|c|c}
    \toprule
         &COMPAS & celebA & MNIST-Invert &MNIST-USPS \\
    \midrule
    Deep SVDD  & 0.97    & 285.10   &   25.73  & 13.12  \\
    \hline
    Ours       & 8.54    & 1703.49   &   167.53  & 231.78  \\
  \bottomrule
\end{tabular}
  \vskip -0.15in
\end{table}

\section{Conclusions and Future Work}
\label{sec:conclusion}
This paper studied the fairness problem of deep anomaly detection methods and proposed a novel deep fair anomaly detection approach (deep fair SVDD). Deep fair SVDD is a method that uses deep neural networks to embed the data into a feature space where the normal data are closely clustered to the centroid. Adversarial training is used so that a discriminatory network cannot predict the protected status.
Further, we propose two measures of the group-level fairness for deep anomaly detection problems. Given the ground truth labels, we can directly measure the \emph{$p \%$ -rule} (equation \ref{eq:fairness}) for the abnormal group. We also propose \emph{distribution distance} (equation \ref{eq:distribution_distance}), which can measure the overall fairness without knowing the labels of anomaly instances. We have conducted extensive empirical studies to evaluate the usefulness of our proposed approach. Firstly, our experiments show that deep anomaly detection methods will generate unfair predictions, even if the training data is balanced with respect to the binary protected state variables. Secondly, we evaluate our proposed deep fair SVDD and compare it to the deep anomaly detection baselines in various data sets. We demonstrate that our proposed work can achieve satisfying fairness results with minimal loss of anomaly detection performance. Next, we analyze the hyper-parameter $\lambda$ which controls the trade-off between fairness and anomaly detection performance within our model and analyze the learned embeddings to study how our proposed model makes fair decisions. 

In this paper, we limited ourselves to studying group-level fairness for deep anomaly detection problems with a single binary protected state variable. We leave for future works to study more complex fair anomaly detection problems such as considering multiple protected state variables, extending to semi-supervised anomaly detection settings (see section \ref{sec:variations}), and improving the training efficiency and scalability.

\bibliographystyle{ACM-Reference-Format}
\bibliography{main}


\end{document}